%% file: main.tex
\documentclass[10pt,twocolumn,letterpaper]{article}

\usepackage[pagenumbers]{cvpr} 

\input{preamble}
\definecolor{cvprblue}{rgb}{0.21,0.49,0.74}
\usepackage[pagebackref,breaklinks,colorlinks,allcolors=cvprblue]{hyperref}
\usepackage{multirow}
\usepackage[table]{xcolor}
\definecolor{maroon}{cmyk}{0,0.87,0.68,0.32}
\usepackage{kotex}
\usepackage{amsmath}
\usepackage{bbm}
\usepackage{wrapfig,lipsum,booktabs}

\def\paperID{****} 
\def\confName{CVPR}
\def\confYear{2026}

\title{Auxiliary Descriptive Knowledge for \\Few-Shot Adaptation of Vision-Language Model}

\author{
SuBeen Lee, GilHan Park, WonJun Moon, Hyun Seok Seong, Jae-Pil Heo\thanks{Corresponding author}\\
Sungkyunkwan University\\
{\tt\small \{leesb7426, a01152a, wjun0830, gustjrdl95, jaepilheo\}@skku.edu}
}

\begin{document}
\maketitle
\input{sec/0_abstract} 
\input{sec/1_introduction}
\input{sec/2_related_work}
\input{sec/3_method}
\input{sec/4_experiments}
\input{sec/5_limitation}
\input{sec/6_conclusion}

{
    \small
    \bibliographystyle{ieeenat_fullname}
    \bibliography{main}
}

\input{supplementary}


\end{document}

%% file: sec/0_abstract.tex
\begin{abstract}
Despite the impressive zero-shot capabilities of Vision-Language Models~(VLMs), they often struggle in downstream tasks with distribution shifts from the pre-training data. 
Few-Shot Adaptation~(FSA-VLM) has emerged as a key solution, typically using Parameter-Efficient Fine-Tuning~(PEFT) to adapt models with minimal data. 
However, these PEFT methods are constrained by their reliance on fixed, handcrafted prompts, which are often insufficient to understand the semantics of classes. 
While some studies have proposed leveraging image-induced prompts to provide additional clues for classification, they introduce prohibitive computational overhead at inference. 
Therefore, we introduce Auxiliary Descriptive Knowledge~(ADK), a novel framework that efficiently enriches text representations without compromising efficiency. 
ADK first leverages a Large Language Model to generate a rich set of descriptive prompts for each class offline. These pre-computed features are then deployed in two ways: (1) as Compositional Knowledge, an averaged representation that provides rich semantics, especially beneficial when class names are ambiguous or unfamiliar to the VLM; and (2) as Instance-Specific Knowledge, where a lightweight, non-parametric attention mechanism dynamically selects the most relevant descriptions for a given image. 
This approach provides two additional types of knowledge alongside the handcrafted prompt, thereby facilitating category distinction across various domains. 
Also, ADK acts as a parameter-free, plug-and-play component that enhances existing PEFT methods. 
Extensive experiments demonstrate that ADK consistently boosts the performance of multiple PEFT baselines, setting a new state-of-the-art across various scenarios.
\end{abstract}

%% file: sec/1_introduction.tex
\section{Introduction}
\label{sec:introduction}
\input{figure/1_figure}
Vision-Language Models~(VLMs), such as CLIP~\cite{CLIP}, which are pre-trained on large-scale image-text data, possess strong generalization capabilities. 
These models have demonstrated significant success across various downstream tasks~\cite{WJHS, CBVA, CRIS, ZOC, CORA}.
Despite this success, VLMs often remain sub-optimal without task-oriented fine-tuning, particularly when the task data encounters a distribution shift from the pre-training data~\cite{CoOp, CoCoOp, RepAdapter, RAdapter}.
While fully fine-tuning the VLM is a straightforward solution, this approach is commonly impractical because it is not always easy to acquire sufficient task-specific data.

To address these challenges, Few-Shot Adaptation of Vision-Language Model~(FSA-VLM) has emerged, which adapts VLM to new tasks with a small amount of data~\cite{CoOp}.
To prevent overfitting from limited data while maintaining the generalization capability of VLM, FSA-VLM primarily adopts Parameter-Efficient Fine-Tuning~(PEFT) techniques~\cite{CoCoOp, MaPLe, 2SFS, CLIP-LoRA, RepAdapter}.
These PEFT methods typically update a small number of parameters for the VLM's backbone, leveraging a fixed, handcrafted class-level prompt~(\eg, ``a photo of a \texttt{<class>}''), as shown in \cref{fig:fig1}~(a).

Although the handcrafted prompt provides general knowledge for classification, we claim that it is often insufficient to optimally distinguish object categories.
Specifically, it becomes significantly easier when the prompt provides additional visual information regarding the corresponding class, as depicted in \cref{fig:fig1}~(b).
Building on a similar observation, prior work, such as CoCoOp~\cite{CoCoOp}, has proposed image-induced context vectors that are concatenated with the handcrafted prompt to encode instance-specific visual context.
However, this approach necessitates online computation of the prompt embedding for each incoming image, which introduces non-trivial latency in large-scale operations, as shown in \cref{fig:fig1}~(c).

In this regard, we propose a novel, Auxiliary Descriptive Knowledge~(ADK), which leverages class-descriptive prompts alongside the handcrafted prompt, as illustrated in \cref{fig:fig1}~(b).
Specifically, we first acquire multiple descriptions for each class using a Large Language Model~(LLM), such as GPT-4~\cite{GPT-4}.
These descriptions represent various visual characteristics of the class, and then we use them as descriptive prompts.
After extracting embeddings for descriptive prompts, we convert them into two types of descriptive knowledge: compositional and instance-specific ones.
The compositional knowledge is computed by averaging descriptive prompts to represent the overall context of the corresponding class.
This compositional class knowledge provides general semantics like the handcrafted prompt, but is particularly useful when the class name provides insufficient semantic cues to distinguish image categories~(\eg, ``DH-82'' or ``DHC-1'' in the FGVC Aircraft dataset).
On the other hand, instance-specific knowledge is conditioned on each image by a non-parametric attention mechanism to selectively aggregate the instance-representative descriptions.
It provides visually grounded textual knowledge, enabling the VLM’s similarity matching to capture fine-grained visual details that are missed by compositional knowledge and handcrafted prompts.
By providing additional visual information of each class with two types of knowledge, the model can easily distinguish image categories, while introducing only marginal time overhead, as shown in \cref{fig:fig1}~(c).
Additionally, ADK is highly compatible with existing FSA-VLM methods as a plug-and-play component, as depicted in \cref{fig:fig1}~(d), since ADK does not require any additional parameters or structural modification.

To summarize our contributions,
\begin{itemize}
\item 
We propose ADK, a novel framework for FSA-VLM that leverages class-descriptive prompts alongside the handcrafted prompt.
These descriptive prompts serve as additional visual information for each class, facilitating the model to solve the task while incurring only marginal computational overhead.


\item 
Our method is plug-and-play, demonstrating high compatibility with existing PEFT techniques for FSA-VLM, as it introduces no additional parameters or structural modifications to the VLM's backbone.

\item 
We verify the effectiveness of ADK through extensive experiments across various benchmarks and scenarios for FSA-VLM, achieving state-of-the-art performance.
\end{itemize}

%% file: figure/1_figure.tex
\begin{figure}
    \centering
    \includegraphics[width=0.48\textwidth]{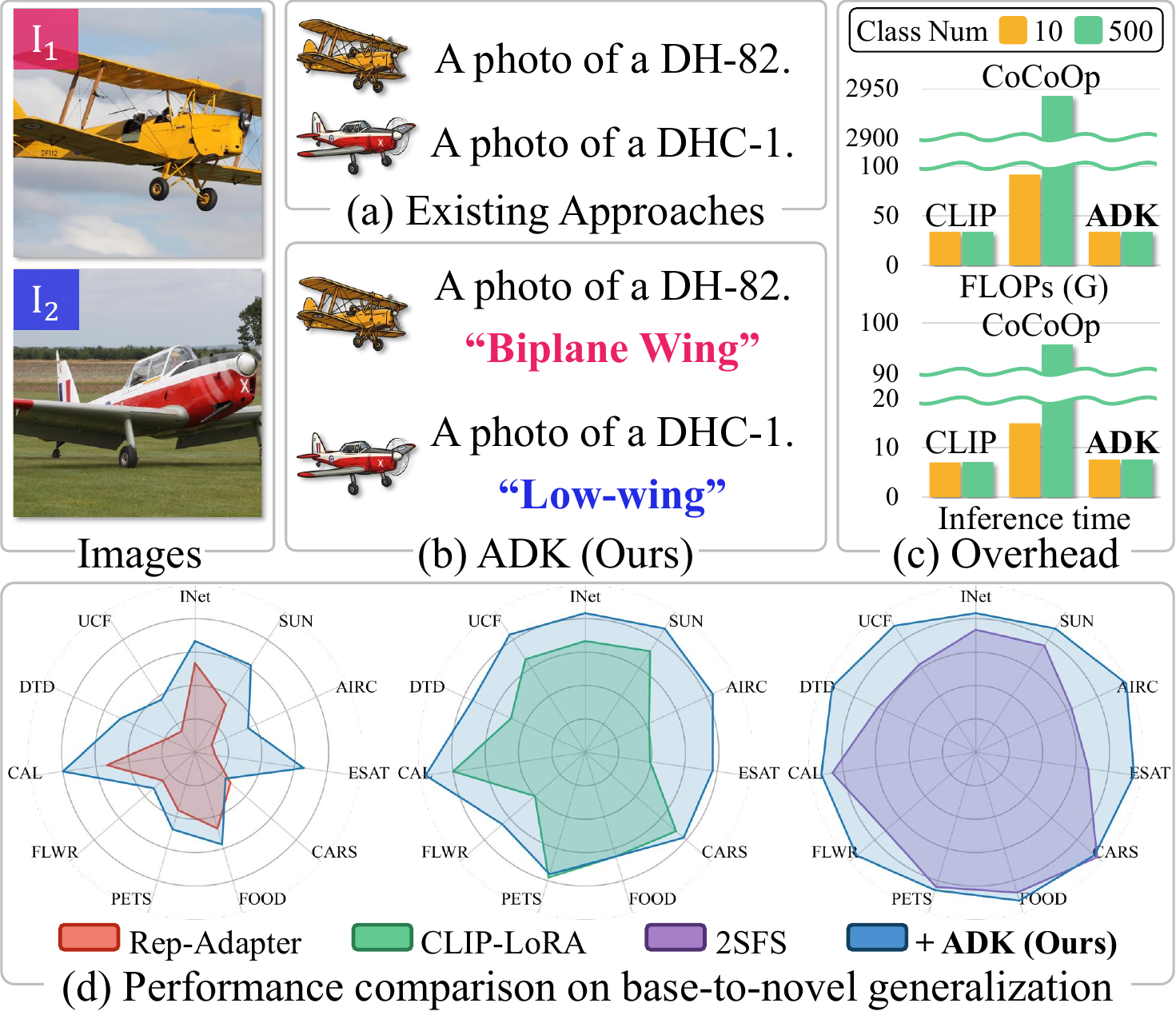}
    \caption{
    \textbf{Comparison between existing approaches and our ADK.}
    (a) Existing approaches rely on a fixed handcrafted prompt~(e.g., ``a photo of \texttt{<class>}''), which often provides insufficient information to distinguish categories, such as the `DH-82'~($I_1$) and the `DHC-1'~($I_2$).
    (b) Our ADK introduces additional class- and image-specific descriptive knowledge~(e.g., ``Biplane Wing'' for $I_1$, ``Low-wing'' for $I_2$). 
    These additional clues enable the model to distinguish categories more effectively.
    (c) Computational cost comparison with CoCoOp, which leverages image-induced context features. 
    In contrast to CoCoOp, which suffers from significant computational overhead, our ADK introduces negligible additional cost. 
    (d) Harmonic mean performance in the base-to-novel scenario when integrating ADK with various methods. 
    ADK demonstrates superior generalization capabilities while preserving adaptation capabilities.
    }
    \label{fig:fig1}
\end{figure}

%% file: sec/2_related_work.tex
\section{Related Work}
\label{sec:related_work}
\subsection{Few-Shot Adaptation Vision-Language Model}
Large-scale pre-trained Vision-Language Models (VLMs), such as CLIP \cite{CLIP}, ALIGN \cite{ALIGN}, LiT~\cite{LIT}, FILIP~\cite{FILIP}, and Florence~\cite{Florence}, have demonstrated impressive zero-shot generalization capabilities. 
Despite this progress, their performance often degrades when applied to downstream tasks that involve domain or distribution shifts~\cite{CoOp, CoCoOp, RepAdapter, RAdapter}.
Furthermore, fine-tuning these models is constrained by the limited availability of labeled data.
To effectively address this degradation in a low-data regime, Few-Shot Adaptation of Vision-Language Model~(FSA-VLM) has emerged as a critical research field.
However, fully fine-tuning the massive VLM parameters using limited data is vulnerable to catastrophic forgetting, thereby leading to performance degradation~\cite{CoOp}.
Therefore, the field has converged on Parameter-Efficient Fine-Tuning~(PEFT), which refines only a small number of the parameters.

\input{figure/2_figure}
\subsection{Parameter-Efficient Fine-Tuning}
PEFT methods can be divided into two main streams: (1) \textit{prompt tuning}, which learns only additional context vectors embedded alongside text or visual tokens~\cite{CoOp, CoCoOp, MaPLe, KgCoOp, ProGrad, PLOT}, and (2) \textit{adapter tuning}, which inserts lightweight trainable modules into the frozen layers of VLM~\cite{RepAdapter, CLIP-LoRA, 2SFS, SuS-X, TaskRes, CLIP-Adapter}.

The concept of prompt tuning was introduced by CoOp~\cite{CoOp} to learn task-specific context vectors, which are prepended to the input text.
These vectors are optimized to convert a handcrafted prompt into a task-specific template. 
CoCoOp~\cite{CoCoOp} extends CoOp by generating those context vectors conditioned on each incoming image, which encode instance-specific visual context.
Addressing the limitations of these uni-modal approaches, MaPLe~\cite{MaPLe} proposes a multi-modal prompt learning that introduces two types of context vectors tuned for the text and visual encoders.

In adapter tuning, CLIP-LoRA~\cite{CLIP-LoRA} applies LoRA~\cite{LoRA} to VLM's attention matrices, which decomposes the large projection weights into smaller, lower-rank matrices to minimize trainable parameters.
RepAdapter~\cite{RepAdapter} employs specialized adapter blocks that are designed using structural reparameterization, where the learned adapter weights are merged with the pretrained VLM.
On the other hand, 2SFS~\cite{2SFS} trains only the layernorm and classifier of VLM using a two-stage protocol.

Despite their effectiveness, both prompt tuning and adapter tuning face limitations.
Prompt tuning requires additional computation in extracting task-specific features for a given image, while adapter tuning typically relies solely on a fixed, handcrafted prompt.
Our proposed framework provides compositional and instance-specific knowledge alongside the handcrafted prompt by utilizing the pre-computed descriptive features.
This pipeline facilitates more effective and efficient adaptation.


%% file: figure/2_figure.tex
\begin{figure*}
    \centering
    \includegraphics[width=0.98\textwidth]{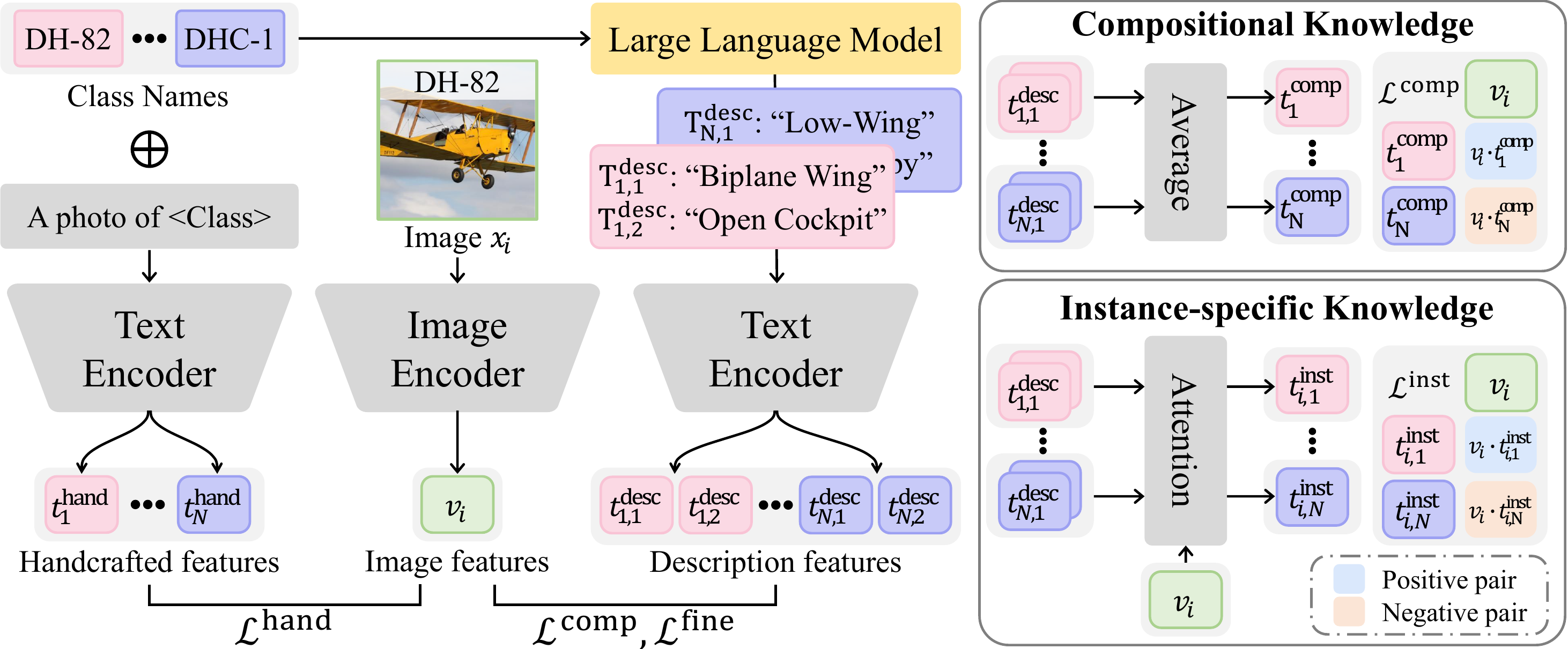}
    \caption{
\textbf{Overall pipeline of Auxiliary Descriptive Knowledge.} 
Given the candidate $N$ classes and an image $x_i$, we query a Large-Language Model to obtain $M$ (here, $M=2$) descriptions $\text{T}^\text{desc}_{n,m}$ for each class, capturing diverse characteristics of that class.
Subsequently, we extract handcrafted features $t^\text{hand}_n$, image features $v_i$, and descriptive features $t^\text{desc}_{n,m}$ for the handcrafted prompt, image, and descriptions.
Then, we generate compositional knowledge $t^\text{comp}_n$ by averaging descriptive features to capture general class-level semantics.
We also derive instance-specific knowledge $t^\text{inst}_{i,n}$ by selectively weighting descriptive features based on their similarity to the image features.
Finally, we utilize these two types of auxiliary knowledge alongside the handcrafted features to optimize the model and predict the image class.
    }
    \label{fig:fig2}
\end{figure*}

%% file: sec/3_method.tex
\section{Method}
\label{sec:method}

\subsection{Overview}
\label{subsec:overview}
The overall pipeline of Auxiliary Descriptive Knowledge~(ADK) framework is illustrated in \cref{fig:fig2}.
Unlike prior works~\cite{CLIP-LoRA, RepAdapter, 2SFS} that rely on a fixed, handcrafted prompt~(\eg, ``a photo of \texttt{<class>}''), ADK incorporates the additional compositional and instance-specific knowledge to facilitate distinguishing categories.
Initially, we query GPT-4~\cite{GPT-4} to gather multiple distinct class-aware descriptions that represent various characteristics of each class.
Based on these descriptions, we generate compositional and instance-specific knowledge for the corresponding class and image.
The compositional knowledge is derived by averaging class-aware descriptions to represent the overall class context.
On the other hand, the instance-specific one captures image-specific characteristics, which are obtained by aggregating descriptions selectively based on their similarity to the input image.
By utilizing these two types of knowledge alongside the handcrafted prompt, we can more effectively distinguish the categories of images.

\subsection{Problem Definition}
\label{subsec:problem_definition}
FSA-VLM aims to adapt pretrained VLM to a specific downstream task in a low-data regime while preserving its generalization ability to novel classes.
Formally, let $\mathcal{D}^{\text{train}}=\left\{\left(x_i, y_i\right) | y_i \in \mathcal{C}^\text{train} \right\}$ denote the sampled target dataset with category set $\mathcal{C}^\text{train}$ containing $N$ classes.
Each class $n \in \mathcal{C}^\text{train}$ contains only $K$ labeled samples, reflecting the few-shot setting.
After adapting VLM with $\mathcal{D}^\text{train}$, we evaluate both adaptation and generalization capabilities of the model using $\mathcal{D}^\text{test} = \left\{\left(x_i,\ y_i\right) | y_i\in \mathcal{C}^\text{test} \right\}$.
When $\mathcal{C}^\text{train} = \mathcal{C}^\text{test}$ (\textit{all-to-all}), the evaluation measures adaptation capabilities, \ie, how well the model adapts to the training classes $\mathcal{C}^\text{train}$. 
On the other hand, in the setting where $\mathcal{C}^\text{test} = \mathcal{C}^\text{train} \cup \mathcal{C}^\text{unseen}$ (\textit{base-to-novel}), it validates generalization capabilities, \ie, how well the model retains its ability to generalize to unseen novel classes $\mathcal{C}^\text{unseen}$ while adapting to base classes $\mathcal{C}^\text{train}$.

\subsection{Class-aware Descriptions}
\label{subsec:class_aware}
The core idea of our ADK is to efficiently integrate additional compositional and instance-specific knowledge alongside the handcrafted prompt.
This strategy enables the model to easily distinguish given images without being overly constrained to a class name, as illustrated in \cref{fig:fig1}~(b).
To avoid the labor-intensive process of manually collecting descriptive knowledge for each class and instance, we leverage the strong descriptive capabilities of GPT-4~\cite{GPT-4}.
We first use GPT-4 to pre-define a vocabulary of class-level common characteristics, which are then combined to generate compositional and instance-specific knowledge.

Specifically, we query GPT-4 with a template prompt (\eg, ``List $M$ sentences to describe the \texttt{<class>}''), to generate $M$ descriptive sentences $\mathcal{T}^\text{desc}_n =\left\{\text{T}^\text{desc}_{n,m} \right\}^M_{m=1}$ for a given $n$-th class.
This class-aware description set $\mathcal{T}^\text{desc}_n$ effectively expands the single class name into various characteristics~(\eg, \texttt{<DH-82>} $\rightarrow$ \texttt{<Biplane Wing>}, \texttt{<Open Cockpit>}), as illustrated in \cref{fig:fig2}.

\subsection{Compositional \& Instance-specific Knowledge}
\label{subsec:instance_aware}
Based on the class-aware descriptions from GPT-4, we derive two types of descriptive knowledge: compositional and image-specific ones.
The compositional knowledge represents an overall visual context of each class, particularly useful when the class name is insufficient to distinguish image categories~(\eg, ``DH-82'' or ``DHC-1'' in \cref{fig:fig2}).
On the other hand, instance-specific knowledge provides a specialized context for each instance that the compositional knowledge or handcrafted prompts overlooked.

Specifically, we first extract the image features $v_i \in \mathbb{R}^D$ of image $x_i$, handcrafted prompt features $t^\text{hand}_n \in \mathbb{R}^D$, and description features $t^\text{desc}_{n,m} \in \mathbb{R}^D$, as follows:
\begin{equation}
    \begin{split}
        \label{eq:feature_extraction}
        v_i &= \Phi^\text{image}\left(x_i\right),
        \\
        t^\text{hand}_n &= \Phi^\text{text}\left(\text{T}^\text{hand}_n\right),
        \\
        t^\text{desc}_{n,m} &= \Phi^\text{text}\left(\text{T}^\text{desc}_{n,m}\right),
    \end{split}
\end{equation}
where $\Phi^\text{image}: I \rightarrow \mathbb{R}^D$ and $\Phi^\text{text}: T \rightarrow \mathbb{R}^D$ denote visual and text encoders of VLM, which map image and text in a visual-text aligned space, respectively.
And, $\text{T}^\text{hand}_n$ is a predefined template~(\eg, ``a photo of \texttt{<class>}'') to generate textual features for each class.

To obtain the compositional class knowledge $t^\text{comp}_n$ that represents the overall context of the $n$-th class, we simply average the descriptive features for that class, as follows: 
\begin{equation}
    \begin{split}
        \label{eq:coarse}
        t^\text{comp}_n = \frac{1}{M} \sum^M_{m=1} t^\text{desc}_{n,m}.
    \end{split}
\end{equation}
On the other hand, the instance-specific knowledge is constructed by a non-parametric attention mechanism based on visual-text similarity to selectively aggregate the descriptive features for each image.
Therefore, the instance-specific knowledge $t^\text{inst}_{i,n}$ for the image features $v_i$ and the $n$-th class is defined as follows:
\begin{equation}
    \begin{split}
        \label{eq:fine_1}
        W_{i,n, m} = \frac{\text{exp}\left(v_i \cdot t^\text{desc}_{n,m} / \tau \right)} {\sum_{m^\prime=1}^{M}\text{exp}\left(v_i \cdot t^\text{desc}_{n,m^\prime}  / \tau \right)} , 
    \end{split}
\end{equation}
\begin{equation}
    \begin{split}
        \label{eq:fine_2}
        t^\text{inst}_{i,n} = \sum_{m=1}^{M} W_{i,n,m} t^\text{desc}_{n,m},
    \end{split}
\end{equation}
where $(\cdot)$ refers to the cosine similarity between two vectors, while $\tau$ is the VLM's logit scale.


These two types of knowledge provide useful information for distinguishing images, in addition to the handcrafted prompt.
Therefore, each knowledge can be utilized to predict the image category, like the handcrafted prompt.
Based on each, the probability for the image $x_i$ to belong to the $n$-th category is formulated as follows:
\begin{equation}
    \begin{split}
        \label{eq:probability_1}
        p\left(y_i=n | t^\text{hand} \right) = \frac{ \text{exp}\left( v_i \cdot t^\text{hand}_{n} / \tau \right) }{ \sum_{n^\prime = 1}^{N} \text{exp}\left( v_i \cdot t^\text{hand}_{n^\prime} / \tau \right) } ,
    \end{split}
\end{equation}
\begin{equation}
    \begin{split}
        \label{eq:probability_2}
        p\left(y_i=n | t^\text{comp} \right) = \frac{ \text{exp}\left( v_i \cdot t^\text{comp}_{n} / \tau \right) }{ \sum_{n^\prime = 1}^{N} \text{exp}\left( v_i \cdot t^\text{comp}_{n^\prime} / \tau \right) } ,
    \end{split}
\end{equation}
\begin{equation}
    \begin{split}
        \label{eq:probability_3}
        p\left(y_i=n | t^\text{inst} \right) = \frac{ \text{exp}\left( v_i \cdot t^\text{inst}_{i,n} / \tau \right) }{ \sum_{n^\prime = 1}^{N} \text{exp}\left( v_i \cdot t^\text{inst}_{i,n^\prime} / \tau \right) } .
    \end{split}
\end{equation}
We define three objective functions for training as follows:
\begin{equation}
    \begin{split}
        \label{eq:loss_1}
        \mathcal{L}^\text{hand} = -\mathbbm{1}_{\left[ y_i=n \right]}\text{log}\left(p\left(y_i=n | t^\text{hand} \right)\right),
    \end{split}
\end{equation}
\begin{equation}
    \begin{split}
        \label{eq:loss_2}
        \mathcal{L}^\text{comp} = -\mathbbm{1}_{\left[ y_i=n \right]}\text{log}\left(p\left(y_i=n | t^\text{comp} \right)\right),
    \end{split}
\end{equation}
\begin{equation}
    \begin{split}
        \label{eq:loss_3}
        \mathcal{L}^\text{inst} = -\mathbbm{1}_{\left[ y_i=n \right]}\text{log}\left(p\left(y_i=n | t^\text{inst} \right)\right),
    \end{split}
\end{equation}
where $\mathbbm{1}_{\left[ y_i=n \right]}$ is the indicator function, which equals 1 if the label of image $y_i$ is $n$ and 0 otherwise.
Then, the final objective function $\mathcal{L}$ is defined as follows:
\begin{equation}
    \begin{split}
        \label{eq:loss}
        \mathcal{L} = \mathcal{L}^\text{hand} + \mathcal{L}^\text{comp} + \mathcal{L}^\text{inst}.
    \end{split}
\end{equation}

During inference, the final prediction $\hat{y}_i$ for the image $x_i$ is computed by combining the predictions from the handcrafted prompt and two types of knowledge, as follows:
\begin{equation}
    \begin{split}
        \label{eq:probability_1}
        \!\!\!\!
        p\!\left(y_i\!=\!n | t^\text{desc} \right) \!=\! 
        \frac{1}{2}\!
        \left(
        p\!\left(y_i\!=\!n | t^\text{comp} \right)
        +
        p\!\left(y_i\!=\!n | t^\text{inst} \right)
        \right)
        ,
    \end{split}
\end{equation}
\begin{equation}
    \begin{split}
        \label{eq:probability_2}
        \hat{y}_i = \underset{n}{\arg\max}
        \left(
        p\left(y_i=n | t^\text{hand}\right)
        +
        p\left(y_i=n | t^\text{desc}\right)
        \right).
    \end{split}
\end{equation}

%% file: sec/4_experiments.tex
\input{table/1_base2novel}
\input{table/2_all2all}

\section{Experiments}
\label{sec:experiments}
In this section, we evaluate our proposed ADK on standard benchmarks for FSA-VLM.
To verify the high compatibility of ADK, we integrate it into various existing methods, including Rep-Adapter~\cite{RepAdapter}, CLIP-LoRA~\cite{CLIP-LoRA}, and 2SFS~\cite{2SFS}. 
For a fair comparison, we reproduce each baseline model, and $\dagger$ in each table indicates the reproduced experiments. 

\vspace{-0.5cm}
\paragraph{Datasets.}
Following 2SFS~\cite{2SFS}, we employ 11 benchmark datasets for FSA-VLM: ImageNet~(INet)~\cite{ImageNet}, SUN397~(SUN)~\cite{SUN397}, FGVC Aircraft~(AIRC)~\cite{FGVC_Aircraft}, EuroSAT~(ESAT)~\cite{EuroSAT}, Stanford Cars~(CARS)~\cite{Stanford_Cars}, Food-101~(FOOD)~\cite{Food-101}, Oxford Pets~(PETS)~\cite{Oxford_Pets}, Oxford Flowers 102~(FLWR)~\cite{Oxford_Flowers}, Caltech101~(CAL)~\cite{Caltech101}, Describable Textures~(DTD)~\cite{DTD}, and UCF-101~(UCF)~\cite{UCF101}.
We follow the dataset split protocol of CoCoOp~\cite{CoCoOp}.

\input{table/3_cd}

\paragraph{Implementation Details.}
For a fair comparison, all baseline models are reproduced following the experimental protocol and two-stage training scheme of our baseline 2SFS~\cite{2SFS}, whether ADK is applied or not.
This scheme involves an initial phase for refining VLM parameters to learn task-level features and a second phase dedicated to tuning the classifier for base classes.
All backbone networks, including ViT-B/16, ViT-B/32, and ViT-L/14, are trained using the AdamW optimizer~\cite{AdamW} with a learning rate of $2\times10^{-4}$, a weight decay of $0.01$, and a mini-batch size of $32$. 
The total number of iterations for the training is fixed to $300 \times K$ for every dataset, where $K$ denotes the number of labeled samples per class.
We adopt the standard handcrafted prompt template~(\eg, ``a photo of \texttt{<class>}''), rather than using dataset-specific templates.
Meanwhile, we acquire $M=20$ class-aware descriptions for each class using GPT-4~\cite{GPT-4}, following the configuration of GPT4Vis~\cite{GPT4Vis}.
Due to overhead from these descriptions, we only use a portion of them corresponding to the classes present in the mini-batch during the first training phase.
All reported results are averaged over three data splits established by CoCoOp~\cite{CoCoOp}.




\subsection{Comparison with state-of-the-art methods}
\label{subsec:comparison_with_state_of_the_art_methods}
Due to space constraints, we primarily present experimental results using the ViT-B/16 backbone architecture in our manuscript.
Results obtained with other architectures, such as ViT-B/32 and ViT-L/14, are provided in the Appendix.

\paragraph{Base-to-Novel.}
This scenario evaluates the adaptation on base categories $\mathcal{C}^\text{train}$ and the generalization to novel classes $\mathcal{C}^\text{unseen}$, as the test set includes both ($\mathcal{C}^\text{test} = \mathcal{C}^\text{train} \cup \mathcal{C}^\text{unseen} $).
This setting is essential for evaluating the key trade-off in FSA-VLM: maintaining the generalization capabilities of VLM while effectively learning new tasks. 
Following standard protocol, we measure performance using the harmonic mean between the accuracies on the base and novel categories.
As shown in \cref{tab:base_to_novel}, ADK consistently improves the overall performances across baselines and achieves new state-of-the-art ($\text{SOTA}$) performances.
These results demonstrate that ADK leverages additional compositional and instance-specific knowledge, enabling the model to better distinguish image categories, whether they are seen or not.
Detailed results for the base and novel categories are provided in the Appendix.

\input{table/4_ablation}
\paragraph{All-to-All.}
In this setup, the training and testing sets are identical ($\mathcal{C}^\text{train} = \mathcal{C}^\text{test}$). 
Therefore, it primarily measures the model's performance on the observed classes during the adaptation.
As shown in \cref{tab:all_to_all}, ADK provides a consistent performance boost to all baseline methods and also records a new SOTA performance.
This demonstrates the superior capability of our approach to effectively fine-tune the VLM on limited data by leveraging compositional and instance-specific knowledge.

\paragraph{Cross-Domain.}
This scenario extends the challenge of generalization by ensuring that the training and testing classes do not overlap ($\mathcal{C}^\text{train} \cap \mathcal{C}^\text{test} = \emptyset$) and belong to fundamentally different domains (or super-classes).
It validates the VLM's generalization capability in the most challenging setting, since knowledge transfer across fundamentally different categories is extremely difficult. 
We train all models using the ImageNet dataset and evaluate them on other domain-specific datasets, utilizing all-to-all splits.
As detailed in \cref{tab:cross-domain}, ADK shows a consistent performance gain compared to baselines.
It implies that integrating additional compositional and image-specific knowledge into the handcrafted prompt enhances the VLM's generalization capability and robustness across entirely new domains.


\subsection{Ablation Study}
\label{subsec:ablation_study}
Unless noted, we utilize 2SFS~\cite{2SFS} for the ablation study.

\paragraph{Component Ablation.}
\cref{tab:ablation_study} presents the performance when our component is applied incrementally. 
(a) shows the standard baseline, where the VLM adaptation relies solely on a fixed, handcrafted prompt. 
(b) denotes the performance when the compositional knowledge is utilized instead of the handcrafted prompt.
Although there seems to be no significant difference between the handcrafted prompt and compositional knowledge, we demonstrate their fundamental difference in a subsequent ablation study.
On the other hand, we observe a performance drop when the model solely relies on the instance-specific knowledge in (c).
This suggests that instance-specific knowledge requires the anchor, which is shared knowledge within the class, to effectively guide the model's adaptation.
(d) shows the effectiveness of the instance-specific knowledge when conjugated with the compositional anchor.
This combination surpasses the baseline, implying that two types of descriptive knowledge provide highly complementary information for object distinction.
Finally, we adopt the auxiliary knowledge alongside the handcrafted prompt in (d) and it shows further enhanced performance, reaching the highest.
This superior performance demonstrates that the handcrafted prompt, compositional, and instance-specific knowledge encode distinct and complementary semantic information.

\input{figure/3_compositional}
\paragraph{Difference between handcrafted prompt and compositional knowledge.}
Although both handcrafted prompt and compositional knowledge encode fixed class-level representations, their utility differs depending on the semantic richness of the class name.
To quantify this difference, we examine how well textual inter-class relationships align with those derived from visual features.
We define three inter-class similarity maps $S\in\mathbb{R}^{|\mathcal{C}^\text{test}| \times |\mathcal{C}^\text{test}|}$ under the pre-trained VLM's feature space, as follows:
\begin{equation}
    \begin{split}
        \label{eq:similarity_map}
        \!\!\!\!\!\!\!
        S^\text{hand} \!=\! t^\text{hand} \!\cdot {t^\text{hand}}^\top
        \!\!,\,\,
        S^\text{comp} \!=\! t^\text{comp} \!\cdot {t^\text{comp}}^\top
        \!\!,\,\,
        S^\text{img} \!=\! \bar{v} \!\cdot {\bar{v}}^\top
        \!\!,
    \end{split}
\end{equation}
where $\bar{v} \in \mathbb{R}^{|\mathcal{C}^\text{test}| \times D}$ represents class-level image prototypes, computed as the average features of images belonging to the same class (\ie, $\bar{v}_n=\sum_{i}\mathbbm{1}_{y_i=n}v_{i} / \sum_{i}\mathbbm{1}_{y_i=n})$.
Then, as shown in \cref{fig:sample} and \cref{tab:kld}, we measure the discrepancy between textual and visual relationships by computing the KL-Divergence~(KLD) of $S^\text{hand}$ and $S^\text{comp}$ with respect to $S^\text{img}$. A larger KLD indicates a misalignment between textual and visual semantics, suggesting that the corresponding textual features fail to capture the general semantics of the image.
In particular, the handcrafted prompt exhibits high KLD and low performance on FGVC Aircraft, since the class name (\eg, `DH-82') provides insufficient semantic information.
In such cases, compositional knowledge serves as a more meaningful and robust semantics.
Conversely, in Stanford Cars, where class names are informative~(\eg, `2006 Ford GT Coupe'), the handcrafted prompt becomes more effective.
For generic datasets such as Caltech101, both handcrafted prompts and compositional knowledge function well, and their relative utility is largely determined by the semantic informativeness of class names.

\paragraph{Visually-grounded Aggregation of Instance-specific Knowledge.}
\input{figure/4_attention}
Instance-specific knowledge is generated by selectively aggregating descriptions according to their similarity to each image, as formulated in \cref{eq:fine_1}.
To verify that this non-parametric mechanism effectively identifies and prioritizes visually-grounded descriptions, we visualize the attention weights $W_{i,n,m}$ for novel classes, as shown in \cref{fig:attention}.
We observe that high weights are assigned to visually-relevant descriptions, confirming that ADK can efficiently construct an accurate, instance-specific description from a predefined set.
More results are in the Appendix.

\input{table/6_overhead}
\paragraph{Computation Overhead in Inference.}
Since our ADK demands additional computation caused by compositional and instance-specific knowledge, we thoroughly analyze its overhead to quantify its efficiency and validate its practical viability.
As shown in \cref{tab:overhead}, ADK introduces only a marginal change in FLOPs and inference time, whereas CoCoOp~\cite{CoCoOp}, which shares a similar motivation for instance-specific context, imposes a severe overhead.
This significant discrepancy stems from their underlying mechanisms.
Specifically, CoCoOp requires the online text encoding with the image-induced context vectors for every image, severely burdening the inference stage.
Conversely, our ADK pre-computes the descriptive features and store them offline, implying that the online process is limited to non-parametric attention for generating instance-specific knowledge and logit computation.
As a result, ADK achieves its goal both effectively and efficiently, adding minimal overhead.

\paragraph{Varying the Number of Class-aware Descriptions.}
\input{table/5_num_descriptions}
Our ADK utilizes $M$ class-aware descriptions generated by GPT-4~\cite{GPT-4}, to establish additional compositional and instance-specific knowledge.
To analyze the impact of description volume, we evaluate ADK under varying values of $M$, as shown in \ref{tab:num_descriptions}.
As $M$ increases from $0$ up to $15$, we observe a rapid performance improvement, demonstrating the substantial benefit of leveraging diverse semantic descriptions. 
However, this trend plateaus; increasing $M$ to 20 yields only marginal gains, indicating that further increases are not significant.
Therefore, we set $M=20$ for all experiments, balancing optimal performance and efficiency.

\input{figure/5_varying_shots}
\paragraph{Varying Shot Configurations.}
While the standard evaluation protocol utilizes a $16$-shot configuration, we further validate ADK's effectiveness under varying $K$-shot settings, from $K=1$ to $K=16$. 
As depicted in \cref{fig:varying_shots}, ADK consistently improves upon the baseline across all $K$-shot configurations. 
These results confirm ADK's robustness and scalability across different few-shot learning regimes.

%% file: table/1_base2novel.tex
\begin{table*}[t]
\setlength{\tabcolsep}{5.8pt} 
\centering
\caption{Performance comparison in the \textit{base-to-novel} setting using the ViT-B/16.
Accuracy is reported as the harmonic mean.}
\label{tab:base_to_novel}
\begin{tabular}{l|c|ccccccccccc}
\toprule
Method & Avg & INet & SUN & AIRC & ESAT & CARS & FOOD & PETS & FLWR & CAL & DTD & UCF \\
\midrule
CLIP~\cite{CLIP} & 71.7 & 70.2 & 72.2 & 31.1 & 60.0 & 68.7 & 90.7 & 94.1 & 74.8 & 95.4 & 56.4 & 73.9 \\
CoOp~\cite{CoOp} & 71.7 & 71.9 & 72.5 & 28.8 & 68.7 & 68.1 & 85.2 & 94.5 & 74.1 & 93.7 & 54.2 & 67.5 \\
CoCoOp~\cite{CoCoOp} & 75.8 & 73.1 & 78.3 & 27.7 & 71.2 & 72.0 & 91.0 & 96.4 & 81.7 & 95.8 & 64.9 & 77.6 \\
MaPLe~\cite{MaPLe} & 78.5 & 73.5 & 79.8 & 36.5 & 82.4 & 73.5 & 91.4 & 96.6 & 82.6 & 96.0 & 68.2 & 80.8 \\
ProGrad~\cite{ProGrad} & 76.2 & 71.5 & 77.6 & 32.8 & 72.7 & 72.9 & 90.0 & 96.3 & 82.0 & 95.9 & 62.5 & 79.4 \\
KgCoOp~\cite{KgCoOp} & 77.0 & 72.8 & 78.4 & 34.8 & 73.5 & 73.4 & 91.1 & 96.2 & 83.7 & 96.0 & 64.4 & 79.7 \\
MMA~\cite{MMA} & 79.9 & 74.0 & 80.4 & 38.3 & 83.9 & 75.7 & 90.7 & 96.7 & 85.5 & 96.2 & 73.4 & 82.2 \\
\midrule
Rep-Adapter$^\dagger$~\cite{RepAdapter} & 77.4 & 73.6 & 78.7 & 36.1 & 71.7 & 75.4 & 88.9 & 93.8 & 82.8 & 95.6 & 69.7 & 79.2\\
\rowcolor{maroon!10} ~~~+ ADK & 78.8 & 74.0 & 80.1 & 38.5 & 79.2 & 75.2 & 89.3 & 94.4 & 83.3 & 96.4 & 72.4 & 81.0 \\
CLIP-LoRA$^\dagger$~\cite{CLIP-LoRA} & 79.4 & 74.0 & 80.6 & 39.2 & 75.5 & 77.6 & 89.6 & 95.9 & 83.8 & 96.4 & 72.4 & 83.3 \\
\rowcolor{maroon!10} ~~~+ ADK & 80.8 & 74.5 & 81.4 & 43.4 & 80.8 & 77.9 & 89.6 & 95.8 & 85.6 & 96.9 & 74.7 & 84.7 \\
2SFS$^\dagger$~\cite{2SFS} & 80.4 & 74.2 & 80.8 & 41.3 & 79.5 & 78.8 & 90.5 & 96.2 & 86.0 & 96.6 & 73.8 & 83.0\\
\rowcolor{maroon!10} ~~~+ ADK & 81.6 & 74.5 & 81.4 & 44.9 & 83.4 & 78.7 & 90.7 & 96.3 & 87.6 & 96.8 & 76.5 & 85.2 \\
\bottomrule
\end{tabular}
\end{table*}

%% file: table/2_all2all.tex
\begin{table*}[t]
\setlength{\tabcolsep}{5.8pt} 
\centering
\caption{
Performance comparison in the \textit{all-to-all} setting using the ViT-B/16.
}
\label{tab:all_to_all}
\begin{tabular}{l|c|ccccccccccc}
\toprule
Method & Avg & INet & SUN & AIRC & ESAT & CARS & FOOD & PETS & FLWR & CAL & DTD & UCF \\
\midrule
CLIP~\cite{CLIP} & 65.1 & 66.7 & 62.6 & 24.7 & 47.5 & 65.3 & 86.1 & 89.1 & 71.4 & 92.9 & 43.6 & 66.7 \\
CoOp~\cite{CoOp} & 80.0 & 71.9 & 74.9 & 43.2 & 85.0 & 82.9 & 84.2 & 92.0 & 96.8 & 95.8 & 69.7 & 83.1 \\
CoCoOp~\cite{CoCoOp} & 75.3 & 71.1 & 72.6 & 33.3 & 73.6 & 72.3 & 87.4 & 93.4 & 89.1 & 95.1 & 63.7 & 77.2 \\
MaPLe~\cite{MaPLe} & 78.6 & 71.9 & 74.5 & 36.8 & 87.5 & 74.3 & 87.4 & 93.2 & 94.2 & 95.4 & 68.4 & 81.4 \\
ProGrad~\cite{ProGrad} & 79.9 & 72.1 & 75.1 & 43.0 & 83.6 & 82.9 & 85.8 & 92.8 & 96.6 & 95.9 & 68.8 & 82.7 \\
KgCoOp~\cite{KgCoOp} & 77.3 & 70.4 & 73.3 & 36.5 & 76.2 & 74.8 & 87.2 & 93.2 & 93.4 & 95.2 & 68.7 & 81.7 \\
MMA~\cite{MMA} & 81.0 & 73.2 & 76.6 & 44.7 & 85.0 & 80.2 & 87.0 & 93.9 & 96.8 & 95.8 & 72.7 & 85.0 \\
\midrule
Rep-Adapter$^\dagger$~\cite{RepAdapter} & 81.2 & 73.4 & 76.3 & 48.5 & 87.5 & 84.6 & 84.6 & 90.5 & 97.7 & 95.8 & 70.6 & 84.2 \\
\rowcolor{maroon!10} ~~~+ ADK & 81.7 & 73.7 & 76.8 & 49.1 & 87.6 & 84.5 & 85.3 & 91.3 & 97.9 & 96.1 & 72.1 & 84.6 \\
CLIP-LoRA$^\dagger$~\cite{CLIP-LoRA} & 83.5 & 73.2 & 76.8 & 54.7 & 92.4 & 86.6 & 85.9 & 92.9 & 98.2 & 96.7 & 73.9 & 87.2 \\
\rowcolor{maroon!10} ~~~+ ADK & 84.0 & 74.0 & 77.3 & 56.2 & 92.5 & 87.4 & 86.2 & 92.9 & 98.3 & 96.8 & 74.6 & 87.5 \\
2SFS$^\dagger$~\cite{2SFS} & 83.1 & 73.7 & 77.4 & 51.0 & 91.9 & 85.8 & 86.5 & 93.0 & 97.8 & 96.6 & 73.8 & 86.7 \\
\rowcolor{maroon!10} ~~~+ ADK & 83.5 & 73.9 & 77.6 & 52.1 & 92.4 & 86.1 & 86.8 & 93.2 & 97.9 & 96.5 & 75.0 & 87.1 \\
\bottomrule
\end{tabular}
\end{table*}

%% file: table/3_cd.tex
\begingroup
\renewcommand{\arraystretch}{1.0} 
\setlength{\tabcolsep}{7.0pt} 
\begin{table*}[t]
\centering
\caption{
Performance comparison in the \textit{cross-domain} setting using the ViT-B/16. Models are trained on ImageNet in the \textit{all-to-all} setting.
}
\label{tab:cross-domain}
\begin{tabular}{l|c|cccccccccc}
\toprule
Method & Avg & SUN & AIRC & ESAT & CARS & FOOD & PETS & FLWR & CAL & DTD & UCF \\
\midrule
CLIP~\cite{CLIP} & 65.0 & 62.6 & 24.7 & 47.5 & 65.3 & 86.1 & 89.1 & 71.4 & 92.9 & 43.6 & 66.7 \\
Rep-Adapter$^\dagger$~\cite{RepAdapter} & 64.3 & 67.1 & 21.4 & 37.6 & 64.0 & 86.0 & 89.9 & 69.9 & 93.9 & 45.4 & 67.9 \\
\rowcolor{maroon!10} ~~~+ ADK & 67.1 & 65.8 & 27.9 & 47.5 & 65.0 & 86.3 & 90.9 & 74.2 & 94.0 & 49.3 & 70.0 \\
CLIP-LoRA$^\dagger$~\cite{CLIP-LoRA} & 64.6 & 67.4 & 24.2 & 34.5 & 65.7 & 86.1 & 89.9 & 70.3 & 93.2 & 45.1 & 69.4 \\
\rowcolor{maroon!10} ~~~+ ADK & 67.4 & 68.3 & 28.4 & 44.4 & 65.5 & 86.2 & 92.0 & 74.9 & 93.6 & 49.4 & 71.6 \\
2SFS$^\dagger$~\cite{2SFS} & 65.1 & 67.3 & 24.6 & 37.9 & 65.5 & 86.2 & 90.2 & 71.7 & 93.6 & 45.7 & 68.6 \\
\rowcolor{maroon!10} ~~~+ ADK & 67.4 & 67.9 & 27.7 & 47.5 & 65.1 & 86.1 & 91.7 & 74.2 & 93.9 & 49.0 & 71.0 \\
\bottomrule
\end{tabular}
\end{table*}
\endgroup

%% file: table/4_ablation.tex
\begingroup
\setlength{\tabcolsep}{4.2pt} 
\renewcommand{\arraystretch}{1.0} 

\begin{table}[t]
\centering
\caption{
Ablation study for each component. 
Hand Prompt, Comp, and Inst denote the handcrafted prompt, compositional knowledge, and instance-specific knowledge, respectively.
B2N and A2A indicate \textit{base-to-novel} and \textit{all-to-all} settings, individually.
Base and Novel represent accuracies in base and novel classes, while HM is their harmonic mean.
}
\label{tab:ablation_study}
\vspace{-0.2cm}
\begin{tabular}{c|c|cc|ccc|c}
\toprule
 & Hand & \multicolumn{2}{c|}{Descriptions} & \multicolumn{3}{c|}{B2N} & \multirow{2}{*}{A2A} \\
\cline{3-7}
 & Prompt & Comp & Inst & Base & Novel & HM & \\
\midrule
(a) & \checkmark & & & 85.5 & 75.8 & 80.4 & 83.1
\\
(b) & & \checkmark & & 85.4 & 75.9 & 80.3 & 83.0
\\
(c) & & & \checkmark & 85.4 & 74.8 & 79.7 & 82.8
\\
(d) & & \checkmark & \checkmark & 85.7 & 76.3 & 80.7 & 83.3
\\
(e) & \checkmark & \checkmark & \checkmark & \textbf{85.9} & \textbf{77.8} & \textbf{81.6} & \textbf{83.5}
\\
\bottomrule
\end{tabular}
\end{table}

\endgroup
\vspace{-0.2cm}

%% file: figure/3_compositional.tex
\begin{figure}[t]
  \centering
  \begin{minipage}[t]{0.48\columnwidth}
    \vspace{0pt}
    \includegraphics[width=\linewidth]{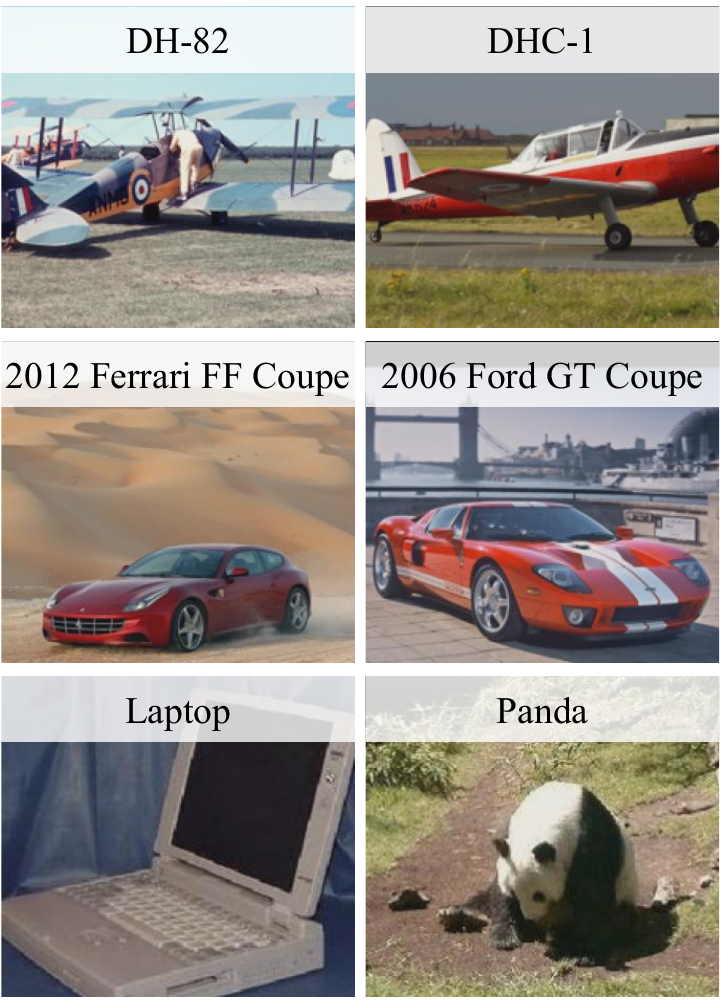}
    \vspace{-0.6cm}
    \caption{Example images and classes for each dataset.
    From top: FGVC Aircraft, Stanford Cars, and Caltech101.}
    \label{fig:sample}
  \end{minipage}\hfill
  \begin{minipage}[t]{0.48\columnwidth}
    \vspace{0pt}
    \centering
    \setlength{\tabcolsep}{3pt}
    \captionof{table}{KL-Divergence (KLD) and Harmonic Mean (HM). 
    Hand and Comp denote handcrafted prompt and compositional knowledge.
    KLD measures the discrepancy between textual and visual inter-class relationships.}
    \label{tab:kld}
    \begin{tabular}{l|c|cc}
      \toprule
       & Data & KLD & HM \\
      \midrule
      Hand & \multirow{2}{*}{AIRC} & 4.19 & 41.3 \\
      Comp &                       & 2.49 & 44.5 \\
      \midrule
      Hand & \multirow{2}{*}{CARS} & 0.70 & 78.8 \\
      Comp &                       & 0.83 & 75.5 \\
      \midrule
      Hand & \multirow{2}{*}{CAL}  & 0.10 & 96.6 \\
      Comp &                       & 0.06 & 96.5 \\
      \bottomrule
    \end{tabular}
  \end{minipage}
\end{figure}

%% file: figure/4_attention.tex
\begin{figure}
    \centering
    \includegraphics[width=0.48\textwidth]{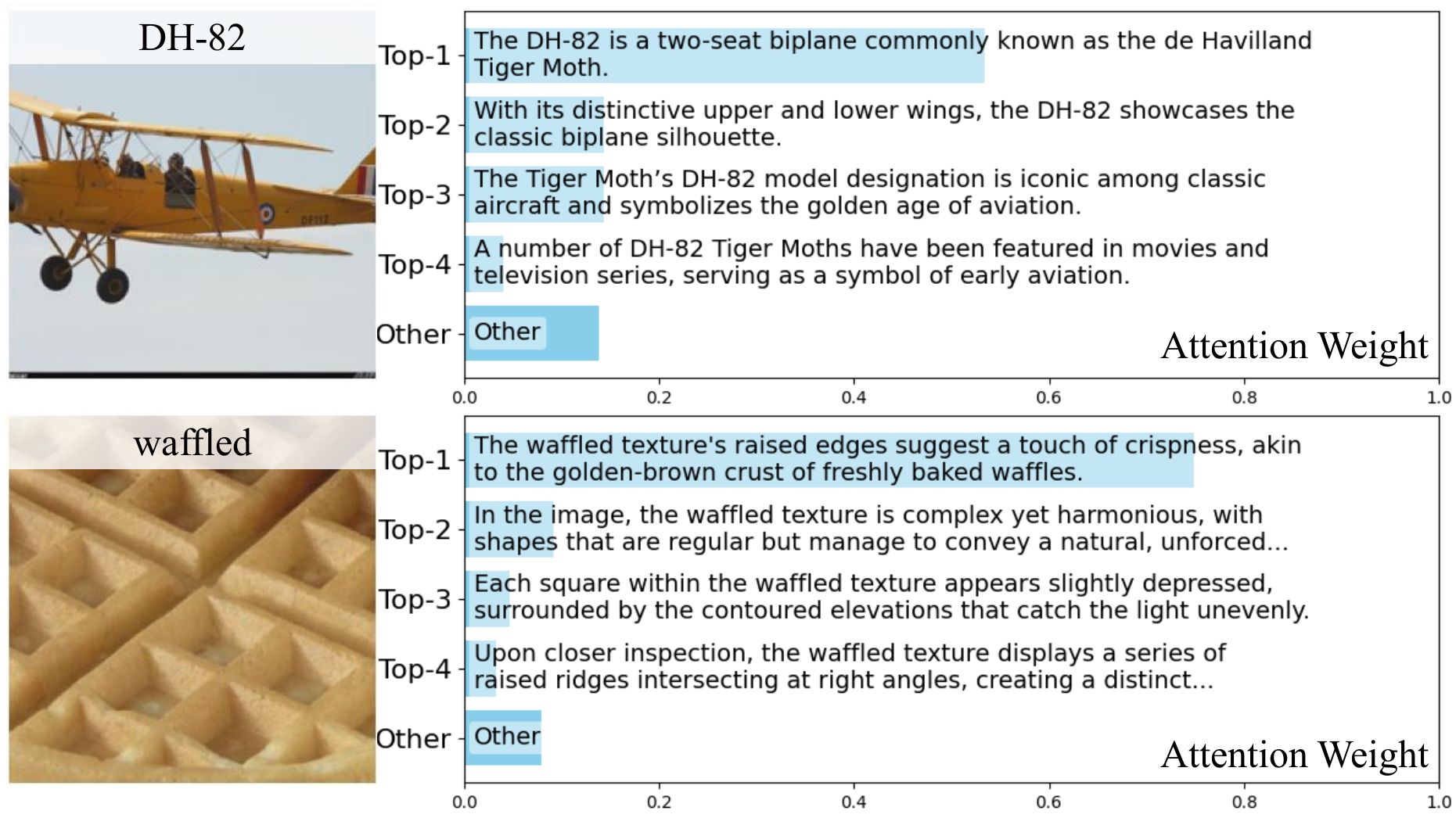}
    \vspace{-0.5cm}
    \caption{
    Visualization of attention weights $W_{i,n,m}$ for instance-specific knowledge, where the image and descriptions belong to the same ground-truth class ($y_i=n$).
    The top four descriptions with high weights are listed in order, with the remaining descriptions grouped under `Other'.
    Best viewed in zoom-in.
    }
    \label{fig:attention}
\end{figure}

%% file: table/6_overhead.tex
\begingroup
\setlength{\tabcolsep}{4.5pt} 
\renewcommand{\arraystretch}{1.0} 
\begin{table}[t]
\centering
\caption{
Computational overhead in inference. 
$N$ is the number of classes, while FLOPs~(Giga-Flops) and Runtime~(ms) indicate computational complexity and inference speed, respectively.
}
\label{tab:overhead}
\vspace{-0.3cm}
\begin{tabular}{l|rr|rr}
\toprule
\multirow{2}{*}{Method} & \multicolumn{2}{c|}{$N$=10}  & \multicolumn{2}{c}{$N$=500} 
\\
 & FLOPs & \multicolumn{1}{c|}{Runtime} & \multicolumn{1}{c}{FLOPs} & \multicolumn{1}{c}{Runtime} \\
\midrule
CLIP~\cite{CLIP} & 33.946 & 7.033 & 33.946 & 7.133
\\
CoCoOp~\cite{CoCoOp} & 92.132 & 14.980 & 2943.246 & 95.778
\\
\rowcolor{maroon!10}ADK & 33.946 & 7.598 & 33.957 & 7.628
\\
\bottomrule
\end{tabular}
\end{table}
\endgroup

%% file: table/5_num_descriptions.tex
\begingroup
\setlength{\tabcolsep}{9pt} 
\renewcommand{\arraystretch}{1.0} 
\begin{table}[t]
\centering
\caption{Performance under varying numbers of descriptions~$M$.
}
\label{tab:num_descriptions}
\vspace{-0.3cm}
\begin{tabular}{c|ccc|c}
\toprule
Number of & \multicolumn{3}{c|}{B2N} & \multirow{2}{*}{A2A} \\
\cline{2-4}
Descriptions & Base & Novel & HM & \\
\midrule
Baseline ($M$=0) & 85.5 & 75.8 & 80.4 & 83.1
\\
$M$=5 & 85.7 & 77.5 & 81.4 & 83.4
\\
$M$=10 & 85.7 & 77.4 & 81.4 & 83.5
\\
$M$=15 & 85.8 & 77.8 & 81.6 & 83.5
\\
$M$=20 & 85.9 & 77.8 & 81.6 & 83.5
\\
\bottomrule
\end{tabular}
\end{table}
\endgroup


%% file: figure/5_varying_shots.tex
\begin{figure}
    \centering
    \includegraphics[width=0.47\textwidth]{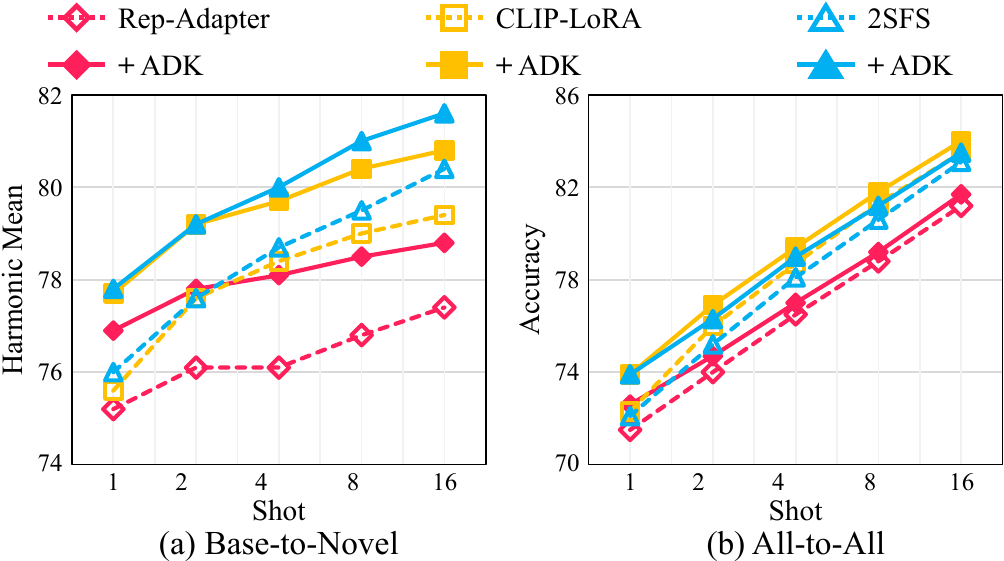}
    \caption{
    K-shot results under varying K.
    Results are averaged on all 11 benchmark datasets for FSA-VLM.
    }
    \label{fig:varying_shots}
\end{figure}

%% file: sec/5_limitation.tex
\section{Limitation and Discussion}
\label{sec:limitation}
Our ADK framework strategically relies on an external LLM to generate distinct class-aware descriptions that represent various characteristics of each class.
This dependency might seem to be a primary limitation of ADK, since the quality of descriptions is not always guaranteed.
However, we note that our core contribution is the efficient and robust ADK pipeline to exploit these descriptions, not the generation of descriptions with LLM. 
Our experimental results demonstrate that the framework is highly effective despite this potential noise. 
We posit two primary reasons for this robustness: (1) compositional knowledge is produced by averaging all description features, effectively mitigating the impact of a few inaccurate outlier descriptions, and (2) instance-specific knowledge naturally assigns higher weights to descriptions that are visually grounded in the input image and lower weights to irrelevant or non-visual text.
Moreover, this dependency implies that as LLMs continue to improve, the performance of the ADK pipeline will be further enhanced.

%% file: sec/6_conclusion.tex
\section{Conclusion}
In this paper, we introduced Auxiliary Descriptive Knowledge~(ADK), a novel framework that efficiently addresses the limitations of handcrafted prompts in Few-Shot Adaptation of Vision-Language Model~(FSA-VLM).
ADK enriches class-aware descriptions for each class, using a Large Language Model.
Utilizing these descriptions, ADK generates compositional and instance-specific knowledge to represent robust class-level semantics and image-relevant details via a non-parametric attention mechanism.
These two types of knowledge can be easily integrated into existing FSA-VLM methods, while incurring negligible overhead, in a plug-and-play manner.
In various scenarios, ADK achieves consistent performance improvements and establishes new state-of-the-art results, demonstrating the effectiveness of integrating auxiliary descriptive knowledge.

%% file: supplementary.tex
\maketitlesupplementary

\section{Additional Experimental Results}
\label{sec:appendix_experiment}
In this section, we present detailed and additional experimental results that were omitted from the main manuscript due to space constraints.

\begin{itemize}
    \item \textbf{$K$-shot Experiments:} Detailed results for the $K$-shot scenario using the ViT-B/16 backbone are shown in \cref{app_tab:tab8} and \cref{app_tab:tab9}.

    \item \textbf{Experiments using various backbone networks:} Results for the \textit{base-to-novel}, \textit{all-to-all}, and \textit{cross-domain} settings—using the ViT-B/16, ViT-B/32, and ViT-L/14 backbones—are presented from \cref{app_tab:tab10} to \cref{app_tab:tab18}.
    The ViT-B/16 results, which partly overlap with those in the main manuscript, are included here to facilitate a convenient comparative analysis across the different backbone networks.
    
    \item \textbf{Additional Visualization:} Furthermore, we offer additional visualizations concerning the attention weights used for constructing the instance-specific knowledge, as shown in \cref{app_fig:fig1}, \cref{app_fig:fig2}, and \cref{app_fig:fig3}.
\end{itemize}

\input{table_supple/11_all2all_kshot}
\input{table_supple/10_base2novel_kshot}

\input{table_supple/1_base2novel_b16}
\input{table_supple/2_base2novel_b32}
\input{table_supple/3_base2novel_l14}

\input{table_supple/4_all2all_b16}
\input{table_supple/5_all2all_b32}
\input{table_supple/6_all2_all_l14}

\input{table_supple/7_cd_b16}
\input{table_supple/8_cd_b32}
\input{table_supple/9_cd_l14}

\input{figure_supple/6_attention}
\input{figure_supple/7_attention}
\input{figure_supple/8_attention}

%% file: table_supple/11_all2all_kshot.tex
\begin{table}[h]
\centering
\caption{Performance of K-shot results under varying K in the \textit{all-to-all} setting, using the ViT-B/16 backbone.}
\label{app_tab:tab8}
\begin{tabular}{l|ccccc}
\toprule
\multirow{2}{*}{Method} & \multicolumn{5}{c}{Number of Shots ($K$)} \\
\cline{2-6}
 & 1 & 2 & 4 & 8 & 16 \\
\midrule
CLIP~\cite{CLIP} & \multicolumn{5}{c}{65.1} \\
Rep-Adapter$^\dagger$~\cite{RepAdapter} & 71.5 & 74.0 & 76.5 & 78.8 & 81.2 \\
\rowcolor{maroon!10} ~~~+ ADK & 72.6 & 74.7 & 77.0 & 79.2 & 81.7 \\
CLIP-LoRA$^\dagger$~\cite{CLIP-LoRA} & 72.3 & 76.0 & 78.7 & 81.3 & 83.5 \\
\rowcolor{maroon!10} ~~~+ ADK & 73.9 & 76.9 & 79.4 & 81.8 & 84.0 \\
2SFS$^\dagger$~\cite{2SFS} & 72.1 & 75.2 & 78.1 & 80.6 & 83.1 \\
\rowcolor{maroon!10} ~~~+ ADK & 73.9 & 76.3 & 79.0 & 81.2 & 83.5 \\
\bottomrule

\end{tabular}
\end{table}

%% file: table_supple/10_base2novel_kshot.tex
\begin{table}[t]
\centering
\caption{Performance of K-shot results under varying K in the \textit{base-to-novel} setting, using the ViT-B/16 backbone.}
\label{app_tab:tab9}
\begin{tabular}{l|ccccc}
\toprule
\multicolumn{6}{c}{\textbf{Accuracy of Base Classes}} \\
\toprule
\multirow{2}{*}{Method} & \multicolumn{5}{c}{Number of Shots ($K$)} \\
\cline{2-6}
 & 1 & 2 & 4 & 8 & 16 \\
\midrule
CLIP~\cite{CLIP} & \multicolumn{5}{c}{69.3} \\
Rep-Adapter$^\dagger$~\cite{RepAdapter} & 75.8 & 78.3 & 79.6 & 81.6 & 83.5 \\
\rowcolor{maroon!10} ~~~+ ADK & 76.8 & 78.6 & 80.2 & 81.9 & 84.1 \\
CLIP-LoRA$^\dagger$~\cite{CLIP-LoRA} & 76.0 & 79.5 & 81.8 & 83.8 & 85.7 \\
\rowcolor{maroon!10} ~~~+ ADK & 77.8 & 80.3 & 82.2 & 84.3 & 86.1 \\
2SFS$^\dagger$~\cite{2SFS} & 76.0 & 79.0 & 81.3 & 83.3 & 85.5 \\
\rowcolor{maroon!10} ~~~+ ADK & 77.8 & 80.1 & 82.1 & 83.9 & 85.9 \\
\bottomrule
\multicolumn{6}{c}{} \\

\toprule
\multicolumn{6}{c}{\textbf{Accuracy of Novel Classes}} \\
\toprule
\multirow{2}{*}{Method} & \multicolumn{5}{c}{Number of Shots ($K$)} \\
\cline{2-6}
 & 1 & 2 & 4 & 8 & 16 \\
\midrule
CLIP~\cite{CLIP} & \multicolumn{5}{c}{74.2} \\
Rep-Adapter$^\dagger$~\cite{RepAdapter} & 74.6 & 74.1 & 72.9 & 72.6 & 72.1 \\
\rowcolor{maroon!10} ~~~+ ADK & 77.1 & 77.0 & 76.1 & 75.4 & 74.2 \\
CLIP-LoRA$^\dagger$~\cite{CLIP-LoRA} & 75.2 & 75.8 & 74.8 & 75.3 & 74.0 \\
\rowcolor{maroon!10} ~~~+ ADK & 77.6 & 78.1 & 77.4 & 76.8 & 76.1 \\
2SFS$^\dagger$~\cite{2SFS} & 76.0 & 76.2 & 76.2 & 76.1 & 75.8 \\
\rowcolor{maroon!10} ~~~+ ADK & 77.7 & 78.3 & 78.1 & 78.2 & 77.8 \\
\bottomrule
\multicolumn{6}{c}{} \\

\toprule
\multicolumn{6}{c}{\textbf{Harmonic Mean between Base and Novel Accuracies}} \\
\toprule
\multirow{2}{*}{Method} & \multicolumn{5}{c}{Number of Shots ($K$)} \\
\cline{2-6}
 & 1 & 2 & 4 & 8 & 16 \\
\midrule
CLIP~\cite{CLIP} & \multicolumn{5}{c}{71.7} \\
Rep-Adapter$^\dagger$~\cite{RepAdapter} & 75.2 & 76.1 & 76.1 & 76.8 & 77.4 \\
\rowcolor{maroon!10} ~~~+ ADK & 76.9 & 77.8 & 78.1 & 78.5 & 78.8 \\
CLIP-LoRA$^\dagger$~\cite{CLIP-LoRA} & 75.6 & 77.6 & 78.4 & 79.0 & 79.4 \\
\rowcolor{maroon!10} ~~~+ ADK & 77.7 & 79.2 & 79.7 & 80.4 & 80.8 \\
2SFS$^\dagger$~\cite{2SFS} & 76.0 & 77.6 & 78.7 & 79.5 & 80.4 \\
\rowcolor{maroon!10} ~~~+ ADK & 77.8 & 79.2 & 80.0 & 81.0 & 81.6 \\
\bottomrule

\end{tabular}
\end{table}

%% file: table_supple/1_base2novel_b16.tex
\begin{table*}[t]
\setlength{\tabcolsep}{5.8pt} 
\centering
\caption{Performance comparison in the \textit{base-to-novel} setting using the ViT-B/16 backbone.}
\label{app_tab:tab10}
\begin{tabular}{l|c|ccccccccccc}
\toprule
\multicolumn{13}{c}{\textbf{Accuracy of Base Classes}} \\
\toprule
Method & Avg & INet & SUN & AIRC & ESAT & CARS & FOOD & PETS & FLWR & CAL & DTD & UCF \\
\midrule
CLIP~\cite{CLIP} & 69.3 & 72.4 & 69.4 & 27.2 & 56.5 & 63.4 & 90.1 & 91.2 & 72.1 & 96.8 & 53.2 & 70.5 \\
CoOp~\cite{CoOp} & 82.7 & 76.5 & 80.6 & 40.4 & 92.2 & 78.1 & 88.3 & 93.7 & 97.6 & 98.0 & 79.4 & 84.7 \\
CoCoOp~\cite{CoCoOp} & 80.5 & 76.0 & 79.7 & 33.4 & 87.5 & 70.5 & 90.7 & 95.2 & 94.9 & 98.0 & 77.0 & 82.3 \\
MaPLe~\cite{MaPLe} & 82.3 & 76.7 & 80.8 & 37.4 & 94.1 & 72.9 & 90.7 & 95.4 & 95.9 & 97.7 & 80.4 & 83.0 \\
ProGrad~\cite{ProGrad} & 82.5 & 77.0 & 81.3 & 40.5 & 90.1 & 77.7 & 90.4 & 95.1 & 95.5 & 98.0 & 77.4 & 84.3 \\
KgCoOp~\cite{KgCoOp} & 80.7 & 75.8 & 80.3 & 36.2 & 85.6 & 71.8 & 90.5 & 94.7 & 95.0 & 97.7 & 77.6 & 82.9 \\
MMA~\cite{MMA} & 83.2 & 77.3 & 82.3 & 40.6 & 85.5 & 78.5 & 90.1 & 95.4 & 97.8 & 98.4 & 83.2 & 86.2 \\
\midrule
Rep-Adapter$^\dagger$~\cite{RepAdapter} & 83.5 & 77.1 & 81.6 & 44.6 & 93.5 & 80.7 & 87.3 & 91.7 & 98.2 & 98.0 & 81.2 & 84.4\\
\rowcolor{maroon!10} ~~~+ ADK & 84.1 & 77.3 & 82.0 & 45.5 & 93.7 & 81.3 & 88.0 & 92.5 & 98.6 & 98.3 & 81.8 & 85.8 \\
CLIP-LoRA$^\dagger$~\cite{CLIP-LoRA} & 85.7 & 77.1 & 82.0 & 51.7 & 96.4 & 83.6 & 88.7 & 94.5 & 98.1 & 98.5 & 84.0 & 88.1 \\
\rowcolor{maroon!10} ~~~+ ADK & 86.1 & 77.8 & 82.5 & 52.4 & 96.3 & 84.4 & 89.0 & 94.3 & 98.3 & 98.6 & 84.9 & 88.9 \\
2SFS$^\dagger$~\cite{2SFS} & 85.5 & 77.8 & 82.8 & 48.4 & 95.6 & 83.0 & 89.7 & 94.6 & 98.4 & 98.5 & 84.0 & 88.0 \\
\rowcolor{maroon!10} ~~~+ ADK & 85.9 & 78.0 & 83.1 & 49.0 & 96.0 & 83.5 & 89.9 & 94.9 & 98.6 & 98.7 & 84.4 & 88.3 \\
\bottomrule
\multicolumn{13}{c}{} \\

\toprule
\multicolumn{13}{c}{\textbf{Accuracy of Novel Classes}} \\
\toprule
Method & Avg & INet & SUN & AIRC & ESAT & CARS & FOOD & PETS & FLWR & CAL & DTD & UCF \\
\midrule
CLIP~\cite{CLIP} & 74.2 & 68.1 & 75.4 & 36.3 & 64.1 & 74.9 & 91.2 & 97.3 & 77.8 & 94.0 & 59.9 & 77.5 \\
CoOp~\cite{CoOp} & 63.2 & 67.9 & 65.9 & 22.3 & 54.7 & 60.4 & 82.3 & 95.3 & 59.7 & 89.8 & 41.2 & 56.1 \\
CoCoOp~\cite{CoCoOp} & 71.7 & 70.4 & 76.9 & 23.7 & 60.0 & 73.6 & 91.3 & 97.7 & 71.8 & 93.8 & 56.0 & 73.5 \\
MaPLe~\cite{MaPLe} & 75.1 & 70.5 & 78.7 & 35.6 & 73.2 & 74.0 & 92.1 & 97.8 & 72.5 & 94.4 & 59.2 & 78.7 \\
ProGrad~\cite{ProGrad} & 70.7 & 66.7 & 74.2 & 27.6 & 60.9 & 68.6 & 89.6 & 97.6 & 71.9 & 93.9 & 52.4 & 74.9 \\
KgCoOp~\cite{KgCoOp} & 73.6 & 70.0 & 76.5 & 33.6 & 64.3 & 75.0 & 91.7 & 97.8 & 74.7 & 94.4 & 55.0 & 76.7 \\
MMA~\cite{MMA} & 76.9 & 71.0 & 78.6 & 36.3 & 82.3 & 73.1 & 91.3 & 98.1 & 75.9 & 94.0 & 65.6 & 80.0 \\
\midrule
Rep-Adapter$^\dagger$~\cite{RepAdapter} & 72.1 & 70.4 & 76.0 & 30.3 & 58.2 & 70.8 & 90.5 & 96.1 & 71.6 & 93.4 & 61.0 & 74.6\\
\rowcolor{maroon!10} ~~~+ ADK & 74.2 & 70.9 & 78.2 & 33.3 & 68.7 & 70.0 & 90.6 & 96.4 & 72.1 & 94.5 & 64.9 & 76.8 \\
CLIP-LoRA$^\dagger$~\cite{CLIP-LoRA} & 74.0 & 71.1 & 79.3 & 31.6 & 62.0 & 72.4 & 90.5 & 97.4 & 73.2 & 94.3 & 63.6 & 79.0 \\
\rowcolor{maroon!10} ~~~+ ADK & 76.1 & 71.5 & 80.4 & 37.0 & 69.6 & 72.3 & 90.3 & 97.2 & 75.7 & 95.2 & 66.7 & 80.9 \\
2SFS$^\dagger$~\cite{2SFS} & 75.8 & 70.9 & 78.9 & 36.0 & 68.1 & 74.9 & 91.4 & 97.8 & 76.3 & 94.7 & 65.7 & 78.6 \\
\rowcolor{maroon!10} ~~~+ ADK & 77.8 & 71.2 & 79.9 & 41.4 & 73.7 & 74.4 & 91.6 & 97.8 & 78.7 & 95.0 & 69.9 & 82.3 \\
\bottomrule
\multicolumn{13}{c}{} \\

\toprule
\multicolumn{13}{c}{\textbf{Harmonic Mean between Base and Novel Accuracies}} \\
\toprule
Method & Avg & INet & SUN & AIRC & ESAT & CARS & FOOD & PETS & FLWR & CAL & DTD & UCF \\
\midrule
CLIP~\cite{CLIP} & 71.7 & 70.2 & 72.2 & 31.1 & 60.0 & 68.7 & 90.7 & 94.1 & 74.8 & 95.4 & 56.4 & 73.9 \\
CoOp~\cite{CoOp} & 71.7 & 71.9 & 72.5 & 28.8 & 68.7 & 68.1 & 85.2 & 94.5 & 74.1 & 93.7 & 54.2 & 67.5 \\
CoCoOp~\cite{CoCoOp} & 75.8 & 73.1 & 78.3 & 27.7 & 71.2 & 72.0 & 91.0 & 96.4 & 81.7 & 95.8 & 64.9 & 77.6 \\
MaPLe~\cite{MaPLe} & 78.5 & 73.5 & 79.8 & 36.5 & 82.4 & 73.5 & 91.4 & 96.6 & 82.6 & 96.0 & 68.2 & 80.8 \\
ProGrad~\cite{ProGrad} & 76.2 & 71.5 & 77.6 & 32.8 & 72.7 & 72.9 & 90.0 & 96.3 & 82.0 & 95.9 & 62.5 & 79.4 \\
KgCoOp~\cite{KgCoOp} & 77.0 & 72.8 & 78.4 & 34.8 & 73.5 & 73.4 & 91.1 & 96.2 & 83.7 & 96.0 & 64.4 & 79.7 \\
MMA~\cite{MMA} & 79.9 & 74.0 & 80.4 & 38.3 & 83.9 & 75.7 & 90.7 & 96.7 & 85.5 & 96.2 & 73.4 & 82.2 \\
\midrule
Rep-Adapter$^\dagger$~\cite{RepAdapter} & 77.4 & 73.6 & 78.7 & 36.1 & 71.7 & 75.4 & 88.9 & 93.8 & 82.8 & 95.6 & 69.7 & 79.2\\
\rowcolor{maroon!10} ~~~+ ADK & 78.8 & 74.0 & 80.1 & 38.5 & 79.2 & 75.2 & 89.3 & 94.4 & 83.3 & 96.4 & 72.4 & 81.0 \\
CLIP-LoRA$^\dagger$~\cite{CLIP-LoRA} & 79.4 & 74.0 & 80.6 & 39.2 & 75.5 & 77.6 & 89.6 & 95.9 & 83.8 & 96.4 & 72.4 & 83.3 \\
\rowcolor{maroon!10} ~~~+ ADK & 80.8 & 74.5 & 81.4 & 43.4 & 80.8 & 77.9 & 89.6 & 95.8 & 85.6 & 96.9 & 74.7 & 84.7 \\
2SFS$^\dagger$~\cite{2SFS} & 80.4 & 74.2 & 80.8 & 41.3 & 79.5 & 78.8 & 90.5 & 96.2 & 86.0 & 96.6 & 73.8 & 83.0\\
\rowcolor{maroon!10} ~~~+ ADK & 81.6 & 74.5 & 81.4 & 44.9 & 83.4 & 78.7 & 90.7 & 96.3 & 87.6 & 96.8 & 76.5 & 85.2 \\
\bottomrule

\end{tabular}
\end{table*}

%% file: table_supple/2_base2novel_b32.tex
\begin{table*}[t]
\setlength{\tabcolsep}{5.8pt} 
\centering
\caption{Performance comparison in the \textit{base-to-novel} setting using the ViT-B/32 backbone.}
\label{app_tab:tab11}
\begin{tabular}{l|c|ccccccccccc}
\toprule
\multicolumn{13}{c}{\textbf{Accuracy of Base Classes}} \\
\toprule
Method & Avg & INet & SUN & AIRC & ESAT & CARS & FOOD & PETS & FLWR & CAL & DTD & UCF \\
\midrule
CLIP~\cite{CLIP} & 67.3 & 67.5 & 69.8 & 21.3 & 55.1 & 60.7 & 85.3 & 90.6 & 72.4 & 94.1 & 54.2 & 69.1 \\
MMA~\cite{MMA} & 78.7 & 72.5 & 80.3 & 31.8 & 71.8 & 73.7 & 85.8 & 93.8 & 95.5 & 97.2 & 79.5 & 83.8 \\
\midrule
Rep-Adapter$^\dagger$~\cite{RepAdapter} & 80.3 & 71.7 & 79.4 & 37.4 & 93.1 & 76.9 & 81.8 & 87.8 & 96.7 & 97.2 & 78.1 & 83.7 \\
\rowcolor{maroon!10} ~~~+ ADK & 80.8 & 72.2 & 80.3 & 37.4 & 93.2 & 77.5 & 82.5 & 88.8 & 96.8 & 97.3 & 78.2 & 84.3 \\
CLIP-LoRA$^\dagger$~\cite{CLIP-LoRA} & 82.3 & 72.1 & 80.3 & 41.3 & 96.2 & 79.0 & 83.8 & 91.7 & 97.4 & 97.7 & 80.4 & 84.9 \\
\rowcolor{maroon!10} ~~~+ ADK & 82.6 & 72.6 & 80.6 & 42.8 & 95.6 & 80.1 & 83.8 & 91.6 & 97.4 & 97.7 & 80.6 & 85.8 \\
2SFS$^\dagger$~\cite{2SFS} & 82.1 & 72.7 & 81.0 & 39.5 & 94.9 & 78.7 & 84.8 & 91.6 & 97.2 & 97.7 & 79.8 & 85.0 \\
\rowcolor{maroon!10} ~~~+ ADK & 82.5 & 73.0 & 81.2 & 40.7 & 95.3 & 78.9 & 85.1 & 92.0 & 97.3 & 97.8 & 80.7 & 85. \\
\bottomrule
\multicolumn{13}{c}{} \\

\toprule
\multicolumn{13}{c}{\textbf{Accuracy of Novel Classes}} \\
\toprule
Method & Avg & INet & SUN & AIRC & ESAT & CARS & FOOD & PETS & FLWR & CAL & DTD & UCF \\
\midrule
CLIP~\cite{CLIP} & 71.7 & 64.1 & 73.0 & 29.3 & 69.8 & 69.7 & 86.9 & 96.9 & 73.7 & 94.0 & 58.2 & 73.0 \\
MMA~\cite{MMA} & 71.0 & 65.8 & 76.6 & 28.7 & 63.0 & 69.3 & 87.1 & 96.3 & 71.6 & 92.6 & 57.0 & 73.5 \\
\midrule
Rep-Adapter$^\dagger$~\cite{RepAdapter} & 67.5 & 66.0 & 74.7 & 23.8 & 58.1 & 64.5 & 85.1 & 94.6 & 62.0 & 92.8 & 52.8 & 68.0 \\
\rowcolor{maroon!10} ~~~+ ADK & 69.4 & 66.9 & 76.5 & 27.7 & 60.6 & 65.3 & 85.4 & 93.9 & 65.3 & 93.4 & 55.9 & 72.5 \\
CLIP-LoRA$^\dagger$~\cite{CLIP-LoRA} & 70.0 & 66.7 & 77.8 & 25.8 & 60.3 & 68.5 & 85.8 & 95.8 & 67.9 & 92.9 & 56.0 & 72.3 \\
\rowcolor{maroon!10} ~~~+ ADK & 71.9 & 67.4 & 78.6 & 31.1 & 65.9 & 67.9 & 85.4 & 94.7 & 70.0 & 93.9 & 61.3 & 74.5 \\
2SFS$^\dagger$~\cite{2SFS} & 71.4 & 66.7 & 77.7 & 30.3 & 63.8 & 70.2 & 87.0 & 95.7 & 70.2 & 93.7 & 55.1 & 74.7 \\
\rowcolor{maroon!10} ~~~+ ADK & 73.9 & 67.2 & 78.4 & 34.5 & 71.5 & 70.5 & 87.5 & 96.6 & 74.8 & 94.4 & 61.6 & 75.8 \\
\bottomrule
\multicolumn{13}{c}{} \\

\toprule
\multicolumn{13}{c}{\textbf{Harmonic Mean between Base and Novel Accuracies}} \\
\toprule
Method & Avg & INet & SUN & AIRC & ESAT & CARS & FOOD & PETS & FLWR & CAL & DTD & UCF \\
\midrule
CLIP~\cite{CLIP} & 69.4 & 65.7 & 71.4 & 24.6 & 61.6 & 64.9 & 86.1 & 93.7 & 73.0 & 94.0 & 56.1 & 71.0 \\
MMA~\cite{MMA} & 74.7 & 69.7 & 78.4 & 30.2 & 67.1 & 71.4 & 86.4 & 95.0 & 81.8 & 94.9 & 66.4 & 78.3 \\
\midrule
Rep-Adapter$^\dagger$~\cite{RepAdapter} & 73.4 & 68.7 & 77.0 & 29.1 & 71.6 & 70.2 & 83.4 & 91.1 & 75.6 & 95.0 & 63.0 & 75.0 \\
\rowcolor{maroon!10} ~~~+ ADK & 74.7 & 69.4 & 78.3 & 31.8 & 73.5 & 70.9 & 83.9 & 91.3 & 78.0 & 95.3 & 65.2 & 78.0 \\
CLIP-LoRA$^\dagger$~\cite{CLIP-LoRA} & 75.6 & 69.3 & 79.0 & 31.7 & 74.2 & 73.4 & 84.8 & 93.7 & 80.0 & 95.2 & 66.0 & 78.1 \\
\rowcolor{maroon!10} ~~~+ ADK & 76.9 & 69.9 & 79.6 & 36.0 & 78.0 & 73.5 & 84.6 & 93.1 & 81.5 & 95.7 & 69.6 & 79.8 \\
2SFS$^\dagger$~\cite{2SFS} & 76.4 & 69.6 & 79.3 & 34.3 & 76.3 & 74.2 & 85.9 & 93.6 & 81.5 & 95.7 & 65.2 & 79.5 \\
\rowcolor{maroon!10} ~~~+ ADK & 78.0 & 70.0 & 79.8 & 37.4 & 81.7 & 74.5 & 86.3 & 94.3 & 84.6 & 96.1 & 69.9 & 80.4 \\
\bottomrule

\end{tabular}
\end{table*}

%% file: table_supple/3_base2novel_l14.tex
\begin{table*}[t]
\setlength{\tabcolsep}{5.8pt} 
\centering
\caption{Performance comparison in the \textit{base-to-novel} setting using the ViT-L/14 backbone.}
\label{app_tab:tab12}
\begin{tabular}{l|c|ccccccccccc}
\toprule
\multicolumn{13}{c}{\textbf{Accuracy of Base Classes}} \\
\toprule
Method & Avg & INet & SUN & AIRC & ESAT & CARS & FOOD & PETS & FLWR & CAL & DTD & UCF \\
\midrule
CLIP~\cite{CLIP} & 76.2 & 79.2 & 73.2 & 37.5 & 70.9 & 74.6 & 93.8 & 93.8 & 80.3 & 95.6 & 59.1 & 79.9 \\
MMA~\cite{MMA} & 85.7 & 83.2 & 85.0 & 50.0 & 77.3 & 85.3 & 94.2 & 96.2 & 99.0 & 98.6 & 85.2 & 88.6 \\
\midrule
Rep-Adapter$^\dagger$~\cite{RepAdapter} & 87.1 & 82.1 & 83.8 & 54.2 & 95.6 & 86.1 & 91.9 & 94.4 & 99.3 & 98.9 & 84.2 & 87.6 \\
\rowcolor{maroon!10} ~~~+ ADK & 87.6 & 82.7 & 84.6 & 55.3 & 95.7 & 86.3 & 92.5 & 95.0 & 99.6 & 99.1 & 84.1 & 88.7 \\
CLIP-LoRA$^\dagger$~\cite{CLIP-LoRA} & 88.9 & 82.6 & 85.4 & 60.3 & 97.2 & 87.8 & 93.3 & 96.1 & 99.4 & 99.1 & 86.7 & 89.6 \\
\rowcolor{maroon!10} ~~~+ ADK & 89.2 & 83.2 & 85.7 & 60.9 & 97.1 & 88.3 & 93.5 & 96.3 & 99.4 & 99.1 & 87.7 & 90.2 \\
2SFS$^\dagger$~\cite{2SFS} & 88.7 & 82.9 & 85.3 & 59.0 & 96.9 & 87.4 & 93.8 & 96.0 & 99.4 & 98.9 & 86.1 & 90.0 \\
\rowcolor{maroon!10} ~~~+ ADK & 88.9 & 83.2 & 85.7 & 59.3 & 96.9 & 87.5 & 93.9 & 96.1 & 99.5 & 98.9 & 86.9 & 90.0 \\
\bottomrule
\multicolumn{13}{c}{} \\

\toprule
\multicolumn{13}{c}{\textbf{Accuracy of Novel Classes}} \\
\toprule
Method & Avg & INet & SUN & AIRC & ESAT & CARS & FOOD & PETS & FLWR & CAL & DTD & UCF \\
\midrule
CLIP~\cite{CLIP} & 80.1 & 74.0 & 77.7 & 44.2 & 82.9 & 84.7 & 94.8 & 96.5 & 83.1 & 95.4 & 67.9 & 79.7 \\
MMA~\cite{MMA} & 79.1 & 76.7 & 81.8 & 42.5 & 62.8 & 83.8 & 95.1 & 98.7 & 80.2 & 96.0 & 70.8 & 81.4 \\
\midrule
Rep-Adapter$^\dagger$~\cite{RepAdapter} & 77.8 & 76.5 & 79.4 & 40.6 & 63.7 & 80.8 & 94.2 & 98.0 & 77.0 & 96.4 & 68.3 & 80.5 \\
\rowcolor{maroon!10} ~~~+ ADK & 80.4 & 77.3 & 81.8 & 45.2 & 74.3 & 81.9 & 94.5 & 97.3 & 81.0 & 97.1 & 70.9 & 83.5 \\
CLIP-LoRA$^\dagger$~\cite{CLIP-LoRA} & 80.0 & 77.2 & 82.2 & 42.3 & 70.6 & 83.2 & 94.6 & 98.5 & 80.2 & 97.5 & 70.9 & 82.8 \\
\rowcolor{maroon!10} ~~~+ ADK & 81.9 & 78.1 & 83.5 & 49.8 & 73.7 & 83.1 & 94.7 & 98.0 & 83.4 & 97.7 & 74.4 & 84.2 \\
2SFS$^\dagger$~\cite{2SFS} & 79.9 & 76.9 & 82.3 & 43.2 & 68.2 & 84.7 & 94.9 & 98.8 & 80.9 & 96.8 & 69.9 & 82.1 \\
\rowcolor{maroon!10} ~~~+ ADK & 82.7 & 77.7 & 83.1 & 51.6 & 75.7 & 84.4 & 95.1 & 98.4 & 85.1 & 98.0 & 75.0 & 85.5 \\
\bottomrule
\multicolumn{13}{c}{} \\

\toprule
\multicolumn{13}{c}{\textbf{Harmonic Mean between Base and Novel Accuracies}} \\
\toprule
Method & Avg & INet & SUN & AIRC & ESAT & CARS & FOOD & PETS & FLWR & CAL & DTD & UCF \\
\midrule
CLIP~\cite{CLIP} & 78.1 & 76.5 & 75.4 & 40.6 & 76.5 & 79.3 & 94.3 & 95.1 & 81.7 & 95.5 & 63.2 & 79.8 \\
MMA~\cite{MMA} & 82.3 & 79.8 & 83.4 & 45.9 & 69.3 & 84.5 & 94.7 & 97.5 & 88.6 & 97.3 & 77.3 & 84.8 \\
\midrule
Rep-Adapter$^\dagger$~\cite{RepAdapter} & 82.2 & 79.2 & 81.6 & 46.4 & 76.5 & 83.4 & 93.1 & 96.2 & 86.7 & 97.6 & 75.4 & 83.9 \\
\rowcolor{maroon!10} ~~~+ ADK & 83.9 & 79.9 & 83.2 & 49.7 & 83.7 & 84.0 & 93.5 & 96.1 & 89.3 & 98.1 & 76.9 & 86.0 \\
CLIP-LoRA$^\dagger$~\cite{CLIP-LoRA} & 84.2 & 79.8 & 83.7 & 49.8 & 81.8 & 85.4 & 93.9 & 97.3 & 88.8 & 98.3 & 78.0 & 86.1 \\
\rowcolor{maroon!10} ~~~+ ADK & 85.4 & 80.6 & 84.6 & 54.8 & 83.8 & 85.7 & 94.1 & 97.1 & 90.7 & 98.4 & 80.5 & 87.1 \\
2SFS$^\dagger$~\cite{2SFS} & 84.1 & 79.8 & 83.8 & 49.9 & 80.0 & 86.0 & 94.4 & 97.4 & 89.2 & 97.8 & 77.2 & 85.8 \\
\rowcolor{maroon!10} ~~~+ ADK & 85.7 & 80.4 & 84.4 & 55.2 & 85.0 & 85.9 & 94.5 & 97.2 & 91.8 & 98.4 & 80.5 & 87.7 \\
\bottomrule

\end{tabular}
\end{table*}

%% file: table_supple/4_all2all_b16.tex
\begin{table*}[t]
\setlength{\tabcolsep}{5.8pt} 
\centering
\caption{
Performance comparison in the \textit{all-to-all} setting using the ViT-B/16 backbone.
}
\label{app_tab:tab13}
\begin{tabular}{l|c|ccccccccccc}
\toprule
Method & Avg & INet & SUN & AIRC & ESAT & CARS & FOOD & PETS & FLWR & CAL & DTD & UCF \\
\midrule
CLIP~\cite{CLIP} & 65.1 & 66.7 & 62.6 & 24.7 & 47.5 & 65.3 & 86.1 & 89.1 & 71.4 & 92.9 & 43.6 & 66.7 \\
CoOp~\cite{CoOp} & 80.0 & 71.9 & 74.9 & 43.2 & 85.0 & 82.9 & 84.2 & 92.0 & 96.8 & 95.8 & 69.7 & 83.1 \\
CoCoOp~\cite{CoCoOp} & 75.3 & 71.1 & 72.6 & 33.3 & 73.6 & 72.3 & 87.4 & 93.4 & 89.1 & 95.1 & 63.7 & 77.2 \\
MaPLe~\cite{MaPLe} & 78.6 & 71.9 & 74.5 & 36.8 & 87.5 & 74.3 & 87.4 & 93.2 & 94.2 & 95.4 & 68.4 & 81.4 \\
ProGrad~\cite{ProGrad} & 79.9 & 72.1 & 75.1 & 43.0 & 83.6 & 82.9 & 85.8 & 92.8 & 96.6 & 95.9 & 68.8 & 82.7 \\
KgCoOp~\cite{KgCoOp} & 77.3 & 70.4 & 73.3 & 36.5 & 76.2 & 74.8 & 87.2 & 93.2 & 93.4 & 95.2 & 68.7 & 81.7 \\
MMA~\cite{MMA} & 81.0 & 73.2 & 76.6 & 44.7 & 85.0 & 80.2 & 87.0 & 93.9 & 96.8 & 95.8 & 72.7 & 85.0 \\
\midrule
Rep-Adapter$^\dagger$~\cite{RepAdapter} & 81.2 & 73.4 & 76.3 & 48.5 & 87.5 & 84.6 & 84.6 & 90.5 & 97.7 & 95.8 & 70.6 & 84.2 \\
\rowcolor{maroon!10} ~~~+ ADK & 81.7 & 73.7 & 76.8 & 49.1 & 87.6 & 84.5 & 85.3 & 91.3 & 97.9 & 96.1 & 72.1 & 84.6 \\
CLIP-LoRA$^\dagger$~\cite{CLIP-LoRA} & 83.5 & 73.2 & 76.8 & 54.7 & 92.4 & 86.6 & 85.9 & 92.9 & 98.2 & 96.7 & 73.9 & 87.2 \\
\rowcolor{maroon!10} ~~~+ ADK & 84.0 & 74.0 & 77.3 & 56.2 & 92.5 & 87.4 & 86.2 & 92.9 & 98.3 & 96.8 & 74.6 & 87.5 \\
2SFS$^\dagger$~\cite{2SFS} & 83.1 & 73.7 & 77.4 & 51.0 & 91.9 & 85.8 & 86.5 & 93.0 & 97.8 & 96.6 & 73.8 & 86.7 \\
\rowcolor{maroon!10} ~~~+ ADK & 83.5 & 73.9 & 77.6 & 52.1 & 92.4 & 86.1 & 86.8 & 93.2 & 97.9 & 96.5 & 75.0 & 87.1 \\
\bottomrule
\end{tabular}
\end{table*}

%% file: table_supple/5_all2all_b32.tex
\begin{table*}[t]
\setlength{\tabcolsep}{5.8pt} 
\centering
\caption{
Performance comparison in the \textit{all-to-all} setting using the ViT-B/32 backbone.
}
\label{app_tab:tab14}
\begin{tabular}{l|c|ccccccccccc}
\toprule
Method & Avg & INet & SUN & AIRC & ESAT & CARS & FOOD & PETS & FLWR & CAL & DTD & UCF \\
\midrule
CLIP~\cite{CLIP} & 61.8 & 61.9 & 62.0 & 19.3 & 45.1 & 60.4 & 80.5 & 87.5 & 67.0 & 91.1 & 42.6 & 62.2 \\
CoOp~\cite{CoOp} & 75.7 & 66.8 & 72.2 & 32.9 & 83.3 & 76.0 & 78.6 & 88.7 & 95.4 & 94.9 & 65.3 & 78.6 \\
CoCoOp~\cite{CoCoOp} & 70.7 & 66.0 & 69.8 & 22.6 & 70.4 & 64.6 & 81.9 & 91.0 & 82.5 & 94.3 & 59.7 & 75.3 \\
MaPLe~\cite{MaPLe} & 74.1 & 66.7 & 72.0 & 28.0 & 83.3 & 66.9 & 82.1 & 91.7 & 89.0 & 95.1 & 63.4 & 77.3 \\
ProGrad~\cite{ProGrad} & 75.9 & 66.9 & 73.2 & 33.3 & 81.0 & 76.1 & 80.1 & 89.3 & 95.1 & 95.0 & 65.8 & 79.6 \\
KgCoOp~\cite{KgCoOp} & 72.1 & 65.4 & 71.0 & 23.7 & 70.1 & 67.3 & 81.7 & 90.8 & 86.1 & 94.4 & 65.1 & 77.5 \\
MMA~\cite{MMA} & 76.6 & 68.0 & 74.0 & 34.0 & 80.1 & 73.5 & 81.4 & 91.5 & 94.3 & 95.6 & 68.9 & 81.7 \\
\midrule
Rep-Adapter$^\dagger$~\cite{RepAdapter} & 77.1 & 68.4 & 74.2 & 38.5 & 86.0 & 78.7 & 78.4 & 86.0 & 96.1 & 94.7 & 66.5 & 80.7 \\
\rowcolor{maroon!10} ~~~+ ADK & 77.7 & 68.6 & 74.8 & 38.9 & 86.2 & 79.2 & 79.3 & 86.9 & 96.4 & 95.2 & 68.1 & 81.3 \\
CLIP-LoRA$^\dagger$~\cite{CLIP-LoRA} & 79.5 & 68.2 & 74.3 & 44.4 & 92.3 & 81.1 & 80.0 & 89.6 & 96.5 & 95.6 & 69.7 & 82.3 \\
\rowcolor{maroon!10} ~~~+ ADK & 80.0 & 69.0 & 74.8 & 45.9 & 92.6 & 82.2 & 80.3 & 89.7 & 96.8 & 95.8 & 69.9 & 83.0 \\
2SFS$^\dagger$~\cite{2SFS} & 79.2 & 68.5 & 74.9 & 41.3 & 91.1 & 80.1 & 81.0 & 89.7 & 96.4 & 95.8 & 70.1 & 82.4 \\
\rowcolor{maroon!10} ~~~+ ADK & 79.7 & 68.8 & 75.1 & 42.4 & 91.6 & 80.6 & 81.3 & 89.9 & 96.9 & 96.0 & 71.0 & 82.7 \\
\bottomrule
\end{tabular}
\end{table*}

%% file: table_supple/6_all2_all_l14.tex
\begin{table*}[t]
\setlength{\tabcolsep}{5.8pt} 
\centering
\caption{
Performance comparison in the \textit{all-to-all} setting using the ViT-L/14 backbone.
}
\label{app_tab:tab15}
\begin{tabular}{l|c|ccccccccccc}
\toprule
Method & Avg & INet & SUN & AIRC & ESAT & CARS & FOOD & PETS & FLWR & CAL & DTD & UCF \\
\midrule
CLIP~\cite{CLIP} & 72.2 & 72.9 & 67.6 & 32.6 & 58.0 & 76.8 & 91.0 & 93.6 & 79.4 & 94.9 & 53.6 & 74.2 \\
CoOp~\cite{CoOp} & 84.6 & 78.2 & 77.5 & 55.2 & 88.3 & 89.0 & 89.8 & 94.6 & 99.1 & 97.2 & 74.4 & 87.3 \\
CoCoOp~\cite{CoCoOp} & 81.7 & 77.8 & 76.7 & 45.2 & 79.8 & 82.7 & 91.9 & 95.4 & 95.3 & 97.4 & 71.4 & 85.2 \\
MaPLe~\cite{MaPLe} & 83.1 & 78.4 & 78.8 & 46.3 & 85.4 & 83.6 & 92.0 & 95.4 & 97.4 & 97.2 & 72.7 & 86.5 \\
ProGrad~\cite{ProGrad} & 84.9 & 78.4 & 78.3 & 55.6 & 89.3 & 88.8 & 90.8 & 94.9 & 98.7 & 97.5 & 73.7 & 87.7 \\
KgCoOp~\cite{KgCoOp} & 82.6 & 76.8 & 76.7 & 47.5 & 83.6 & 83.2 & 91.7 & 95.3 & 96.4 & 97.4 & 73.6 & 86.4 \\
MMA~\cite{MMA} & 84.4 & 79.9 & 80.2 & 56.4 & 76.3 & 88.0 & 92.0 & 95.5 & 98.4 & 97.6 & 75.8 & 88.0 \\
\midrule
Rep-Adapter$^\dagger$~\cite{RepAdapter} & 85.4 & 79.2 & 79.2 & 58.9 & 90.0 & 89.5 & 89.9 & 93.3 & 99.2 & 97.4 & 74.9 & 87.6 \\
\rowcolor{maroon!10} ~~~+ ADK & 85.9 & 79.5 & 80.0 & 59.7 & 90.5 & 89.6 & 90.4 & 93.8 & 99.3 & 97.6 & 75.8 & 88.4 \\
CLIP-LoRA$^\dagger$~\cite{CLIP-LoRA} & 87.3 & 79.0 & 80.2 & 66.2 & 93.0 & 90.7 & 91.0 & 95.1 & 99.3 & 97.7 & 77.4 & 90.4 \\
\rowcolor{maroon!10} ~~~+ ADK & 87.7 & 79.8 & 80.8 & 67.3 & 93.1 & 91.2 & 91.2 & 95.2 & 99.3 & 97.7 & 78.2 & 90.7 \\
2SFS$^\dagger$~\cite{2SFS} & 87.1 & 79.5 & 80.5 & 64.1 & 92.7 & 90.2 & 91.5 & 95.1 & 99.2 & 97.6 & 77.6 & 89.7 \\
\rowcolor{maroon!10} ~~~+ ADK & 87.3 & 79.8 & 80.8 & 64.7 & 92.8 & 90.4 & 91.6 & 95.0 & 99.3 & 97.6 & 78.7 & 90.0 \\
\bottomrule
\end{tabular}
\end{table*}

%% file: table_supple/7_cd_b16.tex
\begingroup
\setlength{\tabcolsep}{7.0pt} 
\renewcommand{\arraystretch}{1.0} 
\begin{table*}[t]
\centering
\caption{
Performance comparison in the \textit{cross-domain} setting using the ViT-B/16.
}
\label{app_tab:tab16}
\begin{tabular}{l|c|cccccccccc}
\toprule
Method & Avg & SUN & AIRC & ESAT & CARS & FOOD & PETS & FLWR & CAL & DTD & UCF \\
\midrule
CLIP~\cite{CLIP} & 65.0 & 62.6 & 24.7 & 47.5 & 65.3 & 86.1 & 89.1 & 71.4 & 92.9 & 43.6 & 66.7 \\
Rep-Adapter$^\dagger$~\cite{RepAdapter} & 64.3 & 67.1 & 21.4 & 37.6 & 64.0 & 86.0 & 89.9 & 69.9 & 93.9 & 45.4 & 67.9 \\
\rowcolor{maroon!10} ~~~+ ADK & 67.1 & 65.8 & 27.9 & 47.5 & 65.0 & 86.3 & 90.9 & 74.2 & 94.0 & 49.3 & 70.0 \\
CLIP-LoRA$^\dagger$~\cite{CLIP-LoRA} & 64.6 & 67.4 & 24.2 & 34.5 & 65.7 & 86.1 & 89.9 & 70.3 & 93.2 & 45.1 & 69.4 \\
\rowcolor{maroon!10} ~~~+ ADK & 67.4 & 68.3 & 28.4 & 44.4 & 65.5 & 86.2 & 92.0 & 74.9 & 93.6 & 49.4 & 71.6 \\
2SFS$^\dagger$~\cite{2SFS} & 65.1 & 67.3 & 24.6 & 37.9 & 65.5 & 86.2 & 90.2 & 71.7 & 93.6 & 45.7 & 68.6 \\
\rowcolor{maroon!10} ~~~+ ADK & 67.4 & 67.9 & 27.7 & 47.5 & 65.1 & 86.1 & 91.7 & 74.2 & 93.9 & 49.0 & 71.0 \\
\bottomrule
\end{tabular}
\end{table*}
\endgroup

%% file: table_supple/8_cd_b32.tex
\begingroup
\setlength{\tabcolsep}{7.0pt} 
\renewcommand{\arraystretch}{1.0} 
\begin{table*}[t]
\centering
\caption{
Performance comparison in the \textit{cross-domain} setting using the ViT-B/32.
}
\label{app_tab:tab17}
\begin{tabular}{l|c|cccccccccc}
\toprule
Method & Avg & SUN & AIRC & ESAT & CARS & FOOD & PETS & FLWR & CAL & DTD & UCF \\
\midrule
CLIP~\cite{CLIP} & 61.8 & 62.0 & 19.3 & 45.1 & 60.4 & 80.5 & 87.5 & 67.0 & 91.1 & 42.6 & 62.2 \\
Rep-Adapter$^\dagger$~\cite{RepAdapter} & 61.2 & 64.7 & 18.2 & 38.6 & 58.2 & 80.2 & 88.1 & 64.6 & 92.3 & 42.4 & 64.5 \\
\rowcolor{maroon!10} ~~~+ ADK & 63.6 & 63.9 & 21.8 & 45.4 & 59.8 & 80.7 & 88.1 & 70.7 & 93.1 & 47.8 & 65.1 \\
CLIP-LoRA$^\dagger$~\cite{CLIP-LoRA} & 61.4 & 65.4 & 18.9 & 36.5 & 59.8 & 80.6 & 87.4 & 66.2 & 91.3 & 43.3 & 64.3 \\
\rowcolor{maroon!10} ~~~+ ADK & 63.7 & 65.9 & 22.3 & 42.8 & 59.4 & 80.9 & 89.0 & 71.2 & 92.7 & 46.4 & 66.9 \\
2SFS$^\dagger$~\cite{2SFS} & 61.7 & 65.3 & 18.6 & 37.8 & 60.1 & 80.5 & 87.8 & 65.9 & 92.3 & 44.3 & 63.9 \\
\rowcolor{maroon!10} ~~~+ ADK & 63.4 & 65.6 & 21.1 & 41.8 & 59.4 & 80.8 & 88.8 & 70.7 & 92.9 & 46.2 & 66.6 \\
\bottomrule
\end{tabular}
\end{table*}
\endgroup

%% file: table_supple/9_cd_l14.tex
\begingroup
\setlength{\tabcolsep}{7.0pt} 
\renewcommand{\arraystretch}{1.0} 
\begin{table*}[t]
\centering
\caption{
Performance comparison in the \textit{cross-domain} setting using the ViT-L/14. 
}
\label{app_tab:tab18}
\begin{tabular}{l|c|cccccccccc}
\toprule
Method & Avg & SUN & AIRC & ESAT & CARS & FOOD & PETS & FLWR & CAL & DTD & UCF \\
\midrule
CLIP~\cite{CLIP} & 72.2 & 67.6 & 32.6 & 58.0 & 76.8 & 91.0 & 93.6 & 79.4 & 94.9 & 53.6 & 74.2 \\
Rep-Adapter$^\dagger$~\cite{RepAdapter} & 71.1 & 70.9 & 30.6 & 50.5 & 75.5 & 91.3 & 93.4 & 75.7 & 96.3 & 51.8 & 75.2 \\
\rowcolor{maroon!10} ~~~+ ADK & 74.4 & 70.3 & 36.5 & 61.8 & 76.7 & 91.4 & 93.2 & 82.0 & 96.1 & 58.7 & 77.6 \\
CLIP-LoRA$^\dagger$~\cite{CLIP-LoRA} & 72.1 & 71.9 & 31.6 & 54.4 & 77.0 & 91.3 & 93.0 & 77.0 & 95.9 & 53.1 & 75.6 \\
\rowcolor{maroon!10} ~~~+ ADK & 74.6 & 72.7 & 38.4 & 59.4 & 76.4 & 91.6 & 94.5 & 81.7 & 96.5 & 56.6 & 78.4 \\
2SFS$^\dagger$~\cite{2SFS} & 72.6 & 71.3 & 31.9 & 55.6 & 76.7 & 91.3 & 94.3 & 78.1 & 96.6 & 54.1 & 75.7 \\
\rowcolor{maroon!10} ~~~+ ADK & 74.6 & 72.3 & 38.5 & 60.2 & 76.9 & 91.5 & 94.4 & 81.2 & 96.6 & 56.8 & 77.8 \\
\bottomrule
\end{tabular}
\end{table*}
\endgroup

%% file: figure_supple/6_attention.tex
\begin{figure*}[!h]
    \centering
    \includegraphics[width=0.98\textwidth]{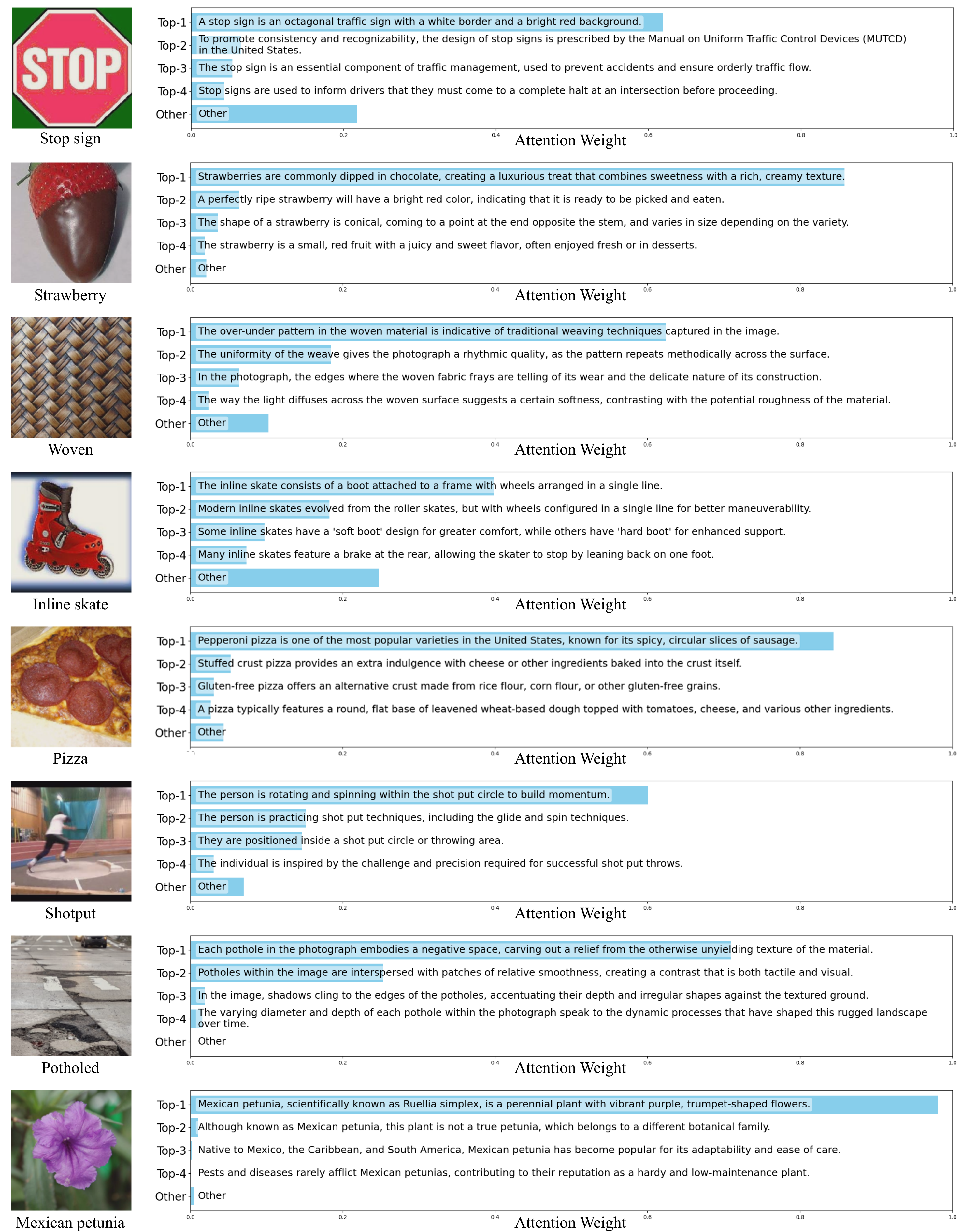}
    \caption{
    Additional Visualization \#1 of attention weight $W_{i,n,m}$ for instance-specific knowledge.
    }
    \label{app_fig:fig1}
\end{figure*}

%% file: figure_supple/7_attention.tex
\begin{figure*}[!h]
    \centering
    \includegraphics[width=0.98\textwidth]{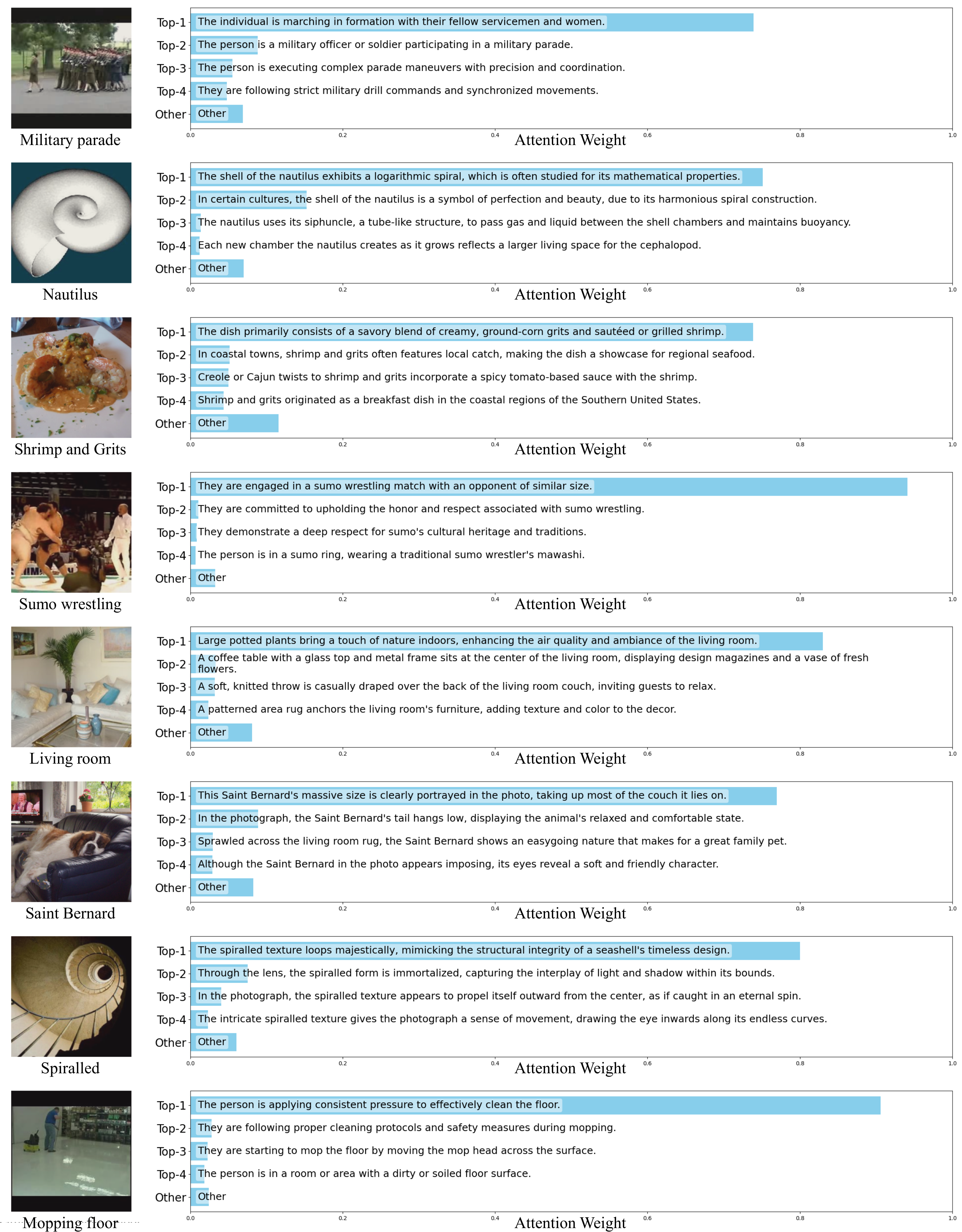}
    \caption{
    Additional Visualization \#2 of attention weight $W_{i,n,m}$ for instance-specific knowledge.
    }
    \label{app_fig:fig2}
\end{figure*}

%% file: figure_supple/8_attention.tex
\begin{figure*}[!h]
    \centering
    \includegraphics[width=0.98\textwidth]{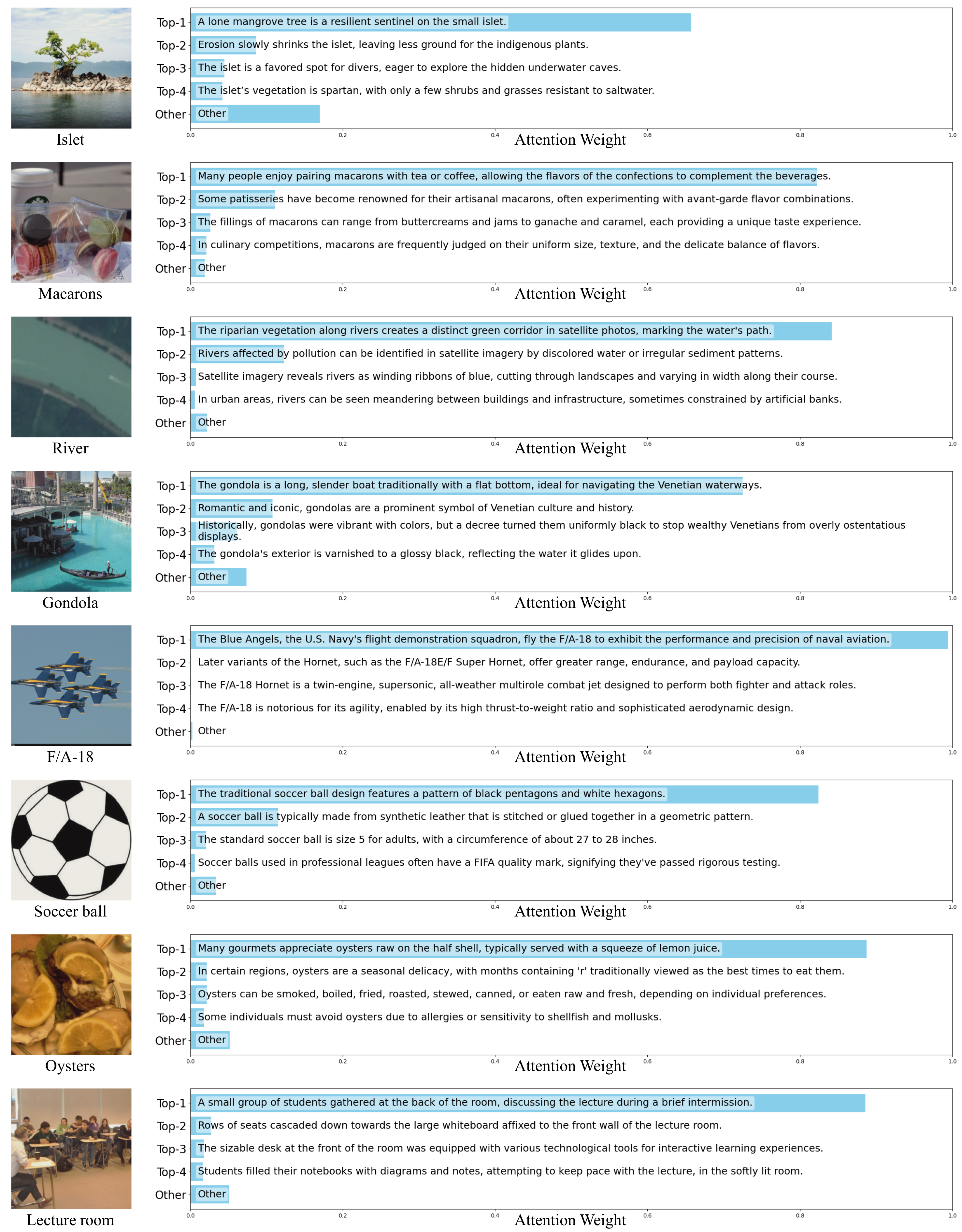}
    \caption{
    Additional Visualization \#3 of attention weight $W_{i,n,m}$ for instance-specific knowledge.
    }
    \label{app_fig:fig3}
\end{figure*}